\documentclass[11pt]{article}

\usepackage[final]{acl}

\usepackage{times}
\usepackage{latexsym}

\usepackage{amsmath}

\usepackage[T1]{fontenc}

\usepackage[utf8]{inputenc}

\usepackage{microtype}

\usepackage{inconsolata}

\usepackage{graphicx}

\usepackage{booktabs}
\usepackage{diagbox}
\usepackage{tabularx} %
\usepackage{listings} %
\usepackage{siunitx}  %

\usepackage{xurl} %

\usepackage{fvextra} %
\usepackage{todonotes}

\makeatletter
\renewcommand\footnotesize{\tiny}
\makeatother

\title{Distilling the Essence:\\ Efficient Reasoning Distillation via Sequence Truncation}

\author{
 \textbf{Wei-Rui Chen\textsuperscript{1,2,*,\S}},
 \textbf{Vignesh Kothapalli\textsuperscript{2}},
 \textbf{Ata Fatahibaarzi\textsuperscript{2}},
 \textbf{Hejian Sang\textsuperscript{2}},
\\
 \textbf{Shao Tang\textsuperscript{2}},
 \textbf{Qingquan Song\textsuperscript{2}},
 \textbf{Zhipeng Wang\textsuperscript{2,*}},
 \textbf{Muhammad Abdul-Mageed \textsuperscript{1,3,*}}
\\
\\
 \textsuperscript{1}The University of British Columbia,
 \textsuperscript{2}LinkedIn,\\
 \textsuperscript{3}Canada Research Chair in NLP and ML
\\
 \small{
   \texttt{weirui.chen@ubc.ca, zhipwang@linkedin.com, muhammad.mageed@ubc.ca}
 }
}

\begin{document}
\maketitle

\begin{abstract}
Distilling the capabilities from a large reasoning model (LRM) to a smaller student model often involves training on substantial amounts of reasoning data. However, knowledge distillation (KD) over lengthy sequences with prompt (P), chain-of-thought (CoT), and answer (A) sections makes the process computationally expensive. In this work, we investigate how the allocation of supervision across different sections (P, CoT, A) affects student performance. Our analysis shows that selective KD over only the CoT tokens can be effective when the prompt and answer information is encompassed by it. Building on this insight, we establish a truncation protocol to quantify computation-quality tradeoffs as a function of sequence length. We observe that beyond a specific length, longer training sequences provide marginal returns for downstream performance but require substantially higher memory and FLOPs. To this end, training on only the first $50\%$ of tokens of every training sequence can retain, on average, $\approx91\%$ of full-sequence performance on math benchmarks while reducing training time, memory usage, and FLOPs by about $50\%$ each. Codes are available at \url{https://github.com/weiruichen01/distilling-the-essence}.

\end{abstract}

\section{Introduction}

Distilling the capabilities of LRMs to smaller counterparts has paved a path for cost-effective, practical deployments in recent years~\citep{hinton2015distilling,hsieh2023distilling,mukherjee2023orcaprogressivelearningcomplex,mitra2023orca2teachingsmall,li-etal-2023-symbolic, guo2025deepseek}. However, datasets curated for this purpose typically comprise long~\emph{chain-of-thought} (CoT) traces~\citep{wei2023chainofthoughtpromptingelicitsreasoning,kojima2023largelanguagemodelszeroshot} with thousands of tokens and impose significant computational overheads during training. To mitigate such issues, we ask if it is possible to extract meaningful distillation signals from truncated training sequences.

Given an input prompt (P), the CoT trace from a capable LRM reflects the multi-step approach to arrive at an answer (A). Essentially, CoT explicitly showcases the model's ability to break down a complex task into sub-problems and its self-reflective phases of corrections. Since smaller models are typically incapable of generating such CoT traces~\cite{wei2022chain}, distilling this ability to them can aid in solving reasoning-intensive tasks in resource-constrained settings~\citep{chernyshev-etal-2025-u, zhang2025realmathcontinuousbenchmarkevaluating, openai2024o1, snell2024scalingllmtesttimecompute, guo2025deepseek}.

In the long-CoT regime, modern LRMs adopt templates that explicitly separate the prompt, a long CoT segment, and the answer~\citep{guo2025deepseek,guha2025openthoughtsdatarecipesreasoning,bespoke_stratos}. Although a growing body of work advocates for the effectiveness of longer CoT traces for reasoning tasks~\cite{chen2025towards}, their utility tends to be task dependent and longer chains do not always translate into higher accuracy~\citep{chen2025think23overthinkingo1like,sui2025stopoverthinkingsurveyefficient,wu2025lessunderstandingchainofthoughtlength,liu-wang-2025-answer, wu2025lessunderstandingchainofthoughtlength}. This raises a fundamental yet under-explored question for distillation: `which segment (prompt, CoT, answer) is the most effective for supervision, and what fraction of the sequence is sufficient to transfer reasoning abilities efficiently?'

In essence, our work addresses the two related research questions:

\begin{itemize}
    \item \textbf{RQ1:} Under controlled experimental conditions, which sections of an example, \emph{prompt (P), CoT,} and \emph{answer (A)}, are actually responsible for transferring reasoning capability to the student?
    \item \textbf{RQ2:} Can training with \emph{early and partial} segments match or approach the performance of training with \emph{full} sequences, thereby reducing training cost without sacrificing accuracy?
    
\end{itemize}

For RQ1, we show that inclusion of the CoT is the dominant factor for effective distillation. In particular, masking (1) the answer segment (P+CoT), (2) the prompt segment (CoT+A), (3) or both (CoT-only) during loss calculation results in minimal performance drop over loss calculation with the full sequence (P+CoT+A). However, masking the CoT segment (in both A-only and P+A settings) results in drastic performance degradation of the student model.

For RQ2, token-budget scaling reveals that beyond a certain length, longer training sequences result in performance plateaus on the downstream math tasks but demand compute resources to scale drastically. In particular, using only the first half of tokens retains $\approx91\%$ of full-sequence performance on average. In contrast, allocating the same budget to the second half of the sequence results in significantly lower accuracy, suggesting that the most critical reasoning signals are concentrated in the early tokens. Collectively, our results position sequence truncation as a new efficiency axis for reasoning distillation, offering a practical path to reduce training time and memory usage. To the best of our knowledge, we are the first to systematically study the supervision of sequence segments and training-time “overthinking” from the perspective of sequence truncation.

\section{Related Work}

\subsection{Long Chain-of-Thought}
A key development in improving Large Language Model (LLM) reasoning is CoT prompting, which encourages models to generate explicit intermediate reasoning steps before giving an answer~\citep{wei2023chainofthoughtpromptingelicitsreasoning}. Such prompting substantially boosts performance on reasoning-intensive tasks in mathematics, science, and code generation~\citep{kojima2023largelanguagemodelszeroshot}. Early CoT work typically produced short, highly compressed rationales. The o1 family of models~\citep{openai2024o1} demonstrated that allocating substantially more inference-time compute can yield large gains on difficult benchmarks, and subsequent test-time scaling studies~\citep{snell2024scalingllmtesttimecompute} systematically verified such benefits. Following \citet{yeo2025demystifyinglongchainofthoughtreasoning}, we refer to such concise chains as \emph{short CoT} and to o1-style, slow-thinking approaches as \emph{long CoT}.

\textit{Short CoT} typically concatenates the reasoning trace with the answer, using a (P, CoT + A) template in which both rationale and answer appear in a single segment with no explicit token-level boundary~\citep{wei2023chainofthoughtpromptingelicitsreasoning,kojima2023largelanguagemodelszeroshot,wang2023selfconsistencyimproveschainthought}. More recent works instead split the response into distinct CoT and answer segments, and formulate a (P, CoT, A) template~\citep{guo2025deepseek, min2024imitateexploreselfimprovereproduction}. For example, DeepSeek-R1 introduces an explicit template in the token stream: the model is trained to place its long CoT inside \verb|<think>|~\dots~\verb|</think>| and the answer afterward~\citep{guo2025deepseek}. STILL-2~\citep{min2024imitateexploreselfimprovereproduction}, OpenThoughts~\citep{guha2025openthoughtsdatarecipesreasoning}, and Bespoke-Stratos~\citep{bespoke_stratos} adopt similar templates that split and place CoT in \verb|<begin_of_thought>|~\dots~\verb|<end_of_thought>|. Such separation allows long reasoning steps to be included in the CoT segment, whereas a concise final solution is placed in the answer (A) segment. Beyond the common practice of utilizing the full sequence (P, CoT, A) for distilling reasoning capabilities to smaller models~\cite{guo2025deepseek}, our work takes a systematic approach to isolating the utility of each of these segments.

\subsection{Distilling Reasoning Signals}\label{subsec:distilling_reasoning_signals}

In the era of LRMs, training and serving full-sized models is extremely compute-intensive. To mitigate such bottlenecks, Knowledge Distillation (KD)~\cite{hinton2015distilling} offers an efficient way to transfer the knowledge of a powerful teacher LRM to a smaller student model~\cite{hsieh2023distilling, mukherjee2023orcaprogressivelearningcomplex, mitra2023orca2teachingsmall}. 
In the \emph{short CoT} regime, supervision from the teacher model is oftentimes emphasized over the response and conditioned on the prompt, as in Orca~\citep{mukherjee2023orcaprogressivelearningcomplex}, Orca2~\citep{mitra2023orca2teachingsmall}, Lion~\citep{jiang-etal-2023-lion}, and MiniLLM~\citep{gu2024minillm}. With some exceptions, prompts are typically conditioned on but are not included in the loss computation.

In the \emph{long CoT} setting, there are two main aspects to consider. The choice of segments for supervision and the choice of reasoning datasets for distillation. For the first aspect, supervision (i.e, loss calculation) can be placed solely on the CoT segment~\citep{shen2025mergeofthoughtdistillation}, on both CoT and answer segments ~\citep{li2025naturalthoughtsselectingdistillingreasoning}, or on the entire sequence (prompt, CoT, and answer triplet)~\citep{xu-etal-2025-twt, xu2025dualheadreasoningdistillationimproving}.
For the second aspect of choosing datasets, we note that several works have challenged the idea that “longer CoT implies better performance” and introduced the notion of reasoning with \emph{overthinking} \citep{chen2025think23overthinkingo1like, sui2025stopoverthinkingsurveyefficient, liu-wang-2025-answer, marjanović2025deepseekr1thoughtologyletsthink}. Utilizing sequences that contain such overthinking signals for distillation can be detrimental to model performance~\citep{wu2025lessunderstandingchainofthoughtlength}.
Prior research on LLM reasoning has also noted that not all parts of a CoT are equally valuable. \citet{marjanović2025deepseekr1thoughtologyletsthink} observe that ``DeepSeek-R1 often undergoes multiple cycles of self-verification, even when it has already arrived at the correct answer.'' This observation suggests that later tokens may not carry significantly useful reasoning signals.
These observations motivate our controlled investigation into whether distillation of early and partial sequences can match or approach the efficacy of full-length sequences. In particular, we focus on computation-quality tradeoffs with increasingly long sequences.

\section{Experiments}

\subsection{Formulation}\label{sec:formulation}
We employ supervised knowledge distillation~\cite{hinton2015distilling, sanh2020distilbertdistilledversionbert}\footnote{For method choice, see Appendix~\ref{appendix:ablation_studies} and \ref{appendix:skd_vs_opkd} for details.} to transfer the capabilities of a large reasoning teacher model, denoted as $ \mathcal{T} $, to a smaller student model that did not undergo any post-training, $ \mathcal{S} $. The training objective for the student model is a combination of a cross-entropy loss against the ground-truth labels and a distillation loss that encourages the student to align with the teacher's output distribution. This composite loss function, $ \mathcal{L} $, for a given input sequence $ x $ from the training corpus, is given by:
$$
\mathcal{L} = \lambda \mathcal{L}_{\text{soft}} + (1 - \lambda) \mathcal{L}_{\text{hard}}
$$
where $ \mathcal{L}_{\text{hard}} $ is the hard label loss, $ \mathcal{L}_{\text{soft}} $ is the soft distillation loss, and $ \lambda \in [0, 1] $ is a hyperparameter that balances the two terms. In our experiments, we set $ \lambda = 0.5 $, same as in \citet{gu2024minillm}, ensuring a balanced contribution from both the soft distillation loss and the hard label loss.\footnote{An ablation of different $\lambda$s is included in Appendix \ref{appendix:ablation_studies} } This approach allows the student to learn not only from the correct labels but also from the nuanced knowledge encapsulated in the teacher's predictions.

The hard loss, $ \mathcal{L}_{\text{hard}} $, is the conventional negative log-likelihood of the ground-truth sequence of $T$ tokens $ y = (y_1, \dots, y_T) $. It is computed as:
$$
\mathcal{L}_{\text{hard}} = - \sum_{t=1}^{T} \log P_{\mathcal{S}}(y_t | y_{<t}, x)
$$
where $ P_{\mathcal{S}}(y_t | y_{<t}, x) $ is the probability assigned by the student model to the true token $ y_t $ at timestep $ t $. The soft loss, $ \mathcal{L}_{\text{soft}} $, uses the forward Kullback-Leibler (KL) divergence to measure the discrepancy between the student's and teacher's output probability distributions. It is defined as the sum of KL divergences over all timesteps:
$$
\mathcal{L}_{\text{soft}} = \sum_{t=1}^{T} D_{\text{KL}}( \sigma(z_{\mathcal{T}}^{(t)}) || \sigma(z_{\mathcal{S}}^{(t)}) )
$$
where $ z_{\mathcal{T}}^{(t)} $ and $ z_{\mathcal{S}}^{(t)} $ are the logit vectors produced by the teacher and student models at timestep $ t $, respectively, and $ \sigma(\cdot) $ is the softmax function.

\subsection{Experimental Setup}

\textbf{Models.}
We select the Qwen3 family~\citep{qwen3technicalreport} because its post-trained models are natively trained with two dedicated tokens, \verb|<think>| and \verb|</think>|, that explicitly delimit the CoT. In contrast to model families that permit reasoning without explicit start/end markers, such as Gemma3~\citep{gemmateam2025gemma3technicalreport} and Mistral~\citep{yentinglin2025_mistral_reasoning}, Qwen3 offers a controlled testbed for comparing masking and truncation strategies. We employ two post-trained teachers (Qwen3-32B and Qwen3-8B) and four base student models (Qwen3-8B, 4B, 1.7B, and 0.6B-Base), which are out-of-the-box pre-trained checkpoints without any post-training. To reflect practical distillation scenarios, we consider the seven teacher-student pairs as shown in Table \ref{tab:qwen3_teacher_student_pairs}.

\begin{table}[htbp]
\centering
\resizebox{\columnwidth}{!}{%
\begin{tabular}{cc@{\hspace{1.5em}}|cc}
\toprule
Teacher & Student & Teacher & Student \\
\midrule
Qwen3-32B & Qwen3-8B-Base   & Qwen3-8B & Qwen3-4B-Base   \\
Qwen3-32B & Qwen3-4B-Base   & Qwen3-8B & Qwen3-1.7B-Base \\
Qwen3-32B & Qwen3-1.7B-Base & Qwen3-8B & Qwen3-0.6B-Base \\
Qwen3-32B & Qwen3-0.6B-Base &          &                 \\
\bottomrule
\end{tabular}%
}
\caption{Seven teacher--student pairs used in our distillation experiments.}
\label{tab:qwen3_teacher_student_pairs}
\end{table}

\textbf{Datasets.} We distill from five reasoning corpora: OpenThoughts-114k\footnote{\url{https://huggingface.co/datasets/open-thoughts/OpenThoughts-114k}} (Openthoughts) \citep{guha2025openthoughtsdatarecipesreasoning}, Bespoke-Stratos-17k\footnote{\url{https://huggingface.co/datasets/bespokelabs/Bespoke-Stratos-17k}} (Bespoke) \citep{bespoke_stratos}, Synthetic-1\footnote{\url{https://huggingface.co/datasets/PrimeIntellect/SYNTHETIC-1}} \citep{2025synthetic1}, Llama-Nemotron-Post-Training-Dataset\footnote{\url{https://huggingface.co/datasets/nvidia/Llama-Nemotron-Post-Training-Dataset}} (Nemotron) \citep{bercovich2025llamanemotronefficientreasoningmodels} and SkyT1-17k\footnote{\url{https://huggingface.co/datasets/NovaSky-AI/Sky-T1_data_17k}} (SkyT1) \citep{sky_t1_2025}. The first four datasets were generated by DeepSeek-R1, whereas the last was produced by QwQ-32B-Preview \citep{qwq-32b-preview}. We use OpenThoughts and Bespoke across all experiments for RQ1 and RQ2. To assess whether the RQ2 results generalize beyond these datasets, we further include Synthetic-1, Nemotron, and SkyT1 exclusively in the RQ2 analysis. Each sequence/example is a linearized dialogue tree: Prompts (P) are user messages; CoT traces and Answers (A) are assistant messages. System messages are removed. Annotation tags \verb|<begin_of_thought>| and \verb|<end_of_thought>| are replaced with \verb|<think>| and \verb|</think>| to match the Qwen3 vocabulary. \verb|<think>| and \verb|</think>| are used to delimit (prompt, CoT, Answer). Token count statistics of each section for the five datasets are listed in Table~\ref{tab:token_count_stats}. Full examples of each dataset and their token distribution of each of the three sections (P, CoT, A) are included in Appendix \ref{appendix:train_dataset_info}.

\begin{table}[htbp]
\centering
\resizebox{\columnwidth}{!}{%
\begin{tabular}{crrrrrrrrrr}
\toprule
&\multicolumn{2}{c}{\textbf{Openthoughts}} &\multicolumn{2}{c}{\textbf{Bespoke}} & \multicolumn{2}{c}{\textbf{Synthetic-1}} & \multicolumn{2}{c}{\textbf{Nemotron}} & \multicolumn{2}{c}{\textbf{SkyT1}}\\
\cmidrule(lr){2-3} \cmidrule(lr){4-5} \cmidrule(lr){6-7} \cmidrule(lr){8-9} \cmidrule(lr){10-11}
Sec & Mean & \% & Mean & \% & Mean & \% & Mean & \% & Mean & \% \\
\midrule
Full & 2497.5  & - & 2165.3 & - & 1854.0 & - & 2437.8 & - & 2277.1 & -\\
P & 116.6 &  4.7 & 120.5 & 5.6 & 100.2 & 5.4 & 87.3 & 3.6 & 139.1 & 6.1 \\
CoT & 1910.4 & 76.6 & 1657.3 & 76.6 & 1402.8 & 75.7 & 2073.3 & 85.1 & 1632.1 & 71.7 \\
A & 468.5 & 18.8 & 385.5 & 17.8 & 349.1 & 18.8 & 275.2 & 11.3 & 503.9 & 22.1\\

\bottomrule
\end{tabular}%
}
\caption{Token count statistics for five training datasets. ``\%'' is the average proportion of tokens in each section relative to the full sequence. %
Distributions are included in Appendix \ref{appendix:train_set_token_count_dist}.}\label{tab:token_count_stats}
\end{table}

To reduce memory usage and ensure that full-length training sequences include all sections (P, CoT, A), we retain only examples with tokenized length below 4k. After filtering, we then sample 40k, 8k, 16k, 16k, and 8k instances from Openthoughts, Bespoke, Synthetic-1, Nemotron, and SkyT1, respectively, for training, and reserve 1k from each corpus for validation.

\textbf{Training.} Across all settings, we train the student models with knowledge distillation, using Fmchisel framework\footnote{\url{https://github.com/linkedin/fmchisel}}~\cite{behdin2025, behdin2025scaling}, for two epochs with AdamW optimizer (learning rate $4\times10^{-6}$, weight decay $0.05$). We use a batch size of 1, with gradient accumulation, across eight Nvidia H100 (80GB) GPUs so the effective batch size
is $8$. See Appendix~\ref{appendix:compute_resources} for detailed compute resource usage. The checkpoint with the lowest validation loss is selected for evaluation.

\textbf{Evaluation.}
We focus our evaluation on mathematical reasoning and consider two widely adopted benchmarks:
AIME24\footnote{\url{https://huggingface.co/datasets/HuggingFaceH4/aime_2024}}~\cite{AIME2024} and AIME25\footnote{\url{https://huggingface.co/datasets/yentinglin/aime_2025}}~\cite{AIME2025}. To form the baselines, we compare student model performance before and after distillation. During evaluation, we set the maximum generation length to $32{,}768$ new tokens, with decoding hyperparameters temperature $0.6$ and top-$p = 0.95$, aligning with the recommended values in Qwen3 Huggingface page\footnote{\url{https://huggingface.co/Qwen/Qwen3-8B}}. We report results with the largest sampling budget (64 samples) for both benchmarks using \texttt{lighteval}\footnote{\url{https://github.com/huggingface/lighteval}}.

\subsection{Methods}

\begin{figure*}[!t]
  \includegraphics[width=1.0\linewidth]{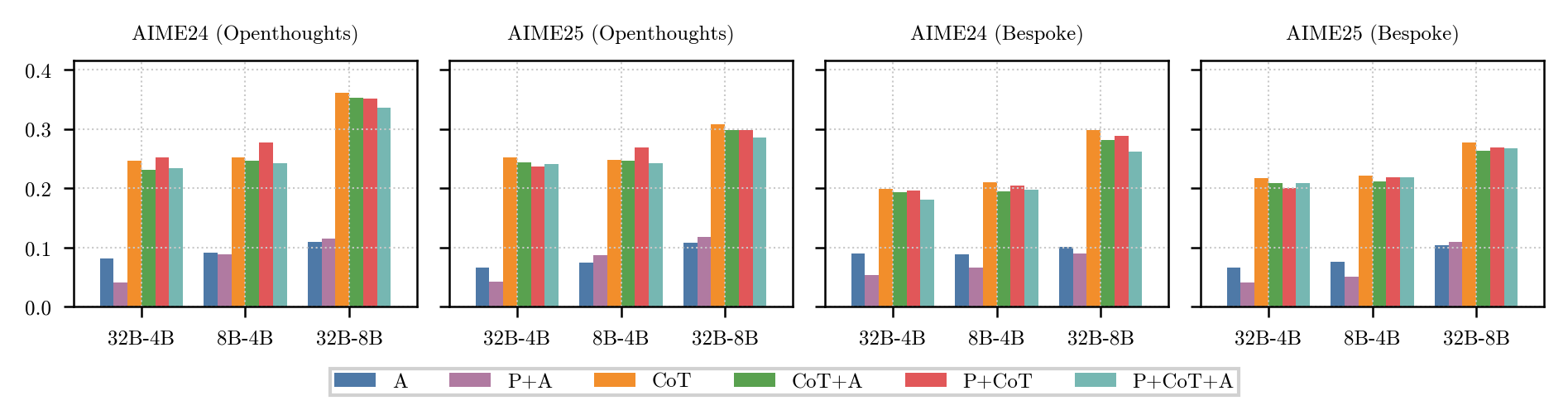}
  \caption {Accuracy on AIME24 and AIME25 under different supervision regions for two different training datasets. Supervising CoT tokens is crucial, with setups that ignore CoT scoring lowest and only modest differences among CoT-inclusive variants.}\label{fig:sec_inclusion_masking}
\end{figure*}

\textbf{Section-wise Supervision (RQ1).} The supervision comprises three sections: P, CoT, and A and we vary which sections are involved in loss calculation. The six conditions are A, P+A, CoT, CoT+A, P+CoT, and P+CoT+A (full-length with all sections included). Following the training setup of previous work \citep{mukherjee2023orcaprogressivelearningcomplex, mitra2023orca2teachingsmall,jiang-etal-2023-lion,gu2024minillm,shi2024instructiontuninglossinstructions,shen2025mergeofthoughtdistillation,li2025naturalthoughtsselectingdistillingreasoning}, we apply a loss mask to all tokens outside the supervised region before computing the objective, while still feeding the entire P+CoT+A sequence through the model in the forward pass. If a section is not included, tokens within that section are conditioned but not involved in loss computation.

\textbf{Early vs. Late Tokens (RQ2).}
To investigate whether useful reasoning signals are concentrated in specific parts of the CoT, we design two complementary experiments targeting compute efficiency.

\emph{Sequence-Length Scaling Behavior.} First, we explore how performance scales with the number of tokens trained. Unlike loss masking, which still incurs forward-pass costs, we examine the performance when tokens are removed entirely. We introduce a dynamic truncation protocol, \emph{Lead-Span Proportion (LSP)}. For each example with tokenized inputs $\mathbf{x}$ and labels $\mathbf{y}$, LSP retains the first $p \in \{0.1, 0.2, \dots, 1.0\}$ fraction of tokens from the concatenation of $\mathbf{x}$ and $\mathbf{y}$, and removes the rightmost $(1-p)$ fraction before training. The truncated tokens, therefore, never enter the model: they are neither seen in the forward pass nor included in the loss. This protocol preserves example-specific variability in kept length, avoids exclusion of answer tokens that can arise under a global max-length cutoff, and retains diversity of sequence lengths.

\emph{Budget Location Ablation.} Second, to determine \emph{where} the token budget should be spent, we fix the training budget at $50\%$ of the tokens (a choice guided by performance saturation, compute efficiency, and experimental symmetry, as detailed in Section \ref{subsec:scaling_behavior}). We compare two conditions:
\emph{Left 50\%}, which keeps the first half of tokens and drops the rest, and \emph{Right 50\%}, which keeps the final half and drops the beginning. This ablation keeps the compute budget fixed while shifting the kept segment of tokens, enabling a clean test of whether early tokens carry richer learning signal than later ones. Better performance in the Left 50\% setting would be consistent with the hypothesis that useful signal concentrates early in the trajectory.

\section{Results and Analysis}

\subsection{Section-wise Supervision (RQ1)}\label{subsec:supervison_diff_sec}

\paragraph{Sub-optimality of full sequence distillation.} As can be seen in Figure~\ref{fig:sec_inclusion_masking}, a consistent pattern emerges across benchmarks and teacher-student pairs: settings that supervise with CoT consistently outperform those that do not. On both benchmarks, answer-only supervision (A) yields the lowest accuracy, and extending the loss to the prompt (P+A) brings only modest gains. Once CoT tokens receive supervision, whether alone (CoT), paired with the answer (CoT+A), paired with the prompt (P+CoT), or combined with both (P+CoT+A), performance improves substantially, with only minor differences among these CoT-inclusive variants.

\begin{figure}[h]
     \centering
  \includegraphics[height=0.25\textheight,keepaspectratio]
  {images/one_training_ex_wide.png}
  \caption{A training example. The text is continuous but divided into three sections to illustrate the idea of Prompt (P), CoT traces (CoT), and Answer (A). We observe that the information in P and A is often covered within CoT. Some intermediate text in CoT has been omitted here for readability. The full example is provided as Figure \ref{fig:complete_training_ex_Abe} in Appendix \ref{appendix:train_dataset_info}.}\label{fig:training_ex_omitted_annotated}
\end{figure}
Overall, the results indicate that supervising the rationale tokens is crucial, while spreading loss to the prompt and answer offers only modest, sometimes negligible, additional benefits. Thus, full-token supervision does not reliably outperform more selective objectives that focus on the reasoning process itself.

\paragraph{CoT implicitly encompasses the prompt and answer sections.} To understand the limited gains from supervising P and A, we manually examined the data and observed that the CoT often encapsulates information from both the prompt and the answer as illustrated in Figure~\ref{fig:training_ex_omitted_annotated}. Typically, the CoT begins by paraphrasing the question to contextualize the problem, then proceeds with reflection-based reasoning steps before deriving the final answer. The answer section typically summarizes the CoT and reiterates the final answer near the end.

\begin{figure*}[!t]
  \includegraphics[width=1.0\linewidth]{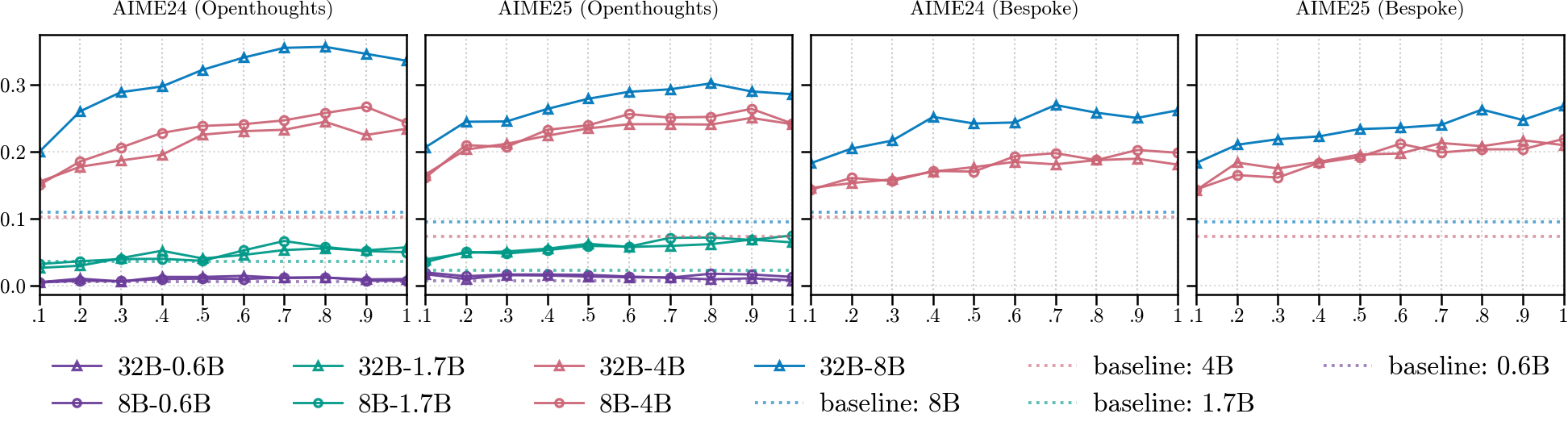}
  \caption {Sequence-length scaling behavior of accuracy against different Lead-Span Proportions (LSP) shows that performance is not linear. The vast majority of performance can be retained when training with only a portion of tokens for each training example. Since student models smaller than 4B performed near baseline on OpenThoughts, we restricted our experiments on Bespoke to student models larger than or equal to 4B.}\label{fig:combined_LSP}
\end{figure*}

\paragraph{Linguistic analysis.} To verify whether P and A are commonly covered by the CoT in our training datasets, we conduct a linguistic analysis on 100 randomly sampled training examples using GPT-5.1 as a judge. We prompt the model to verify that: (1) all substantive information in P and A is already present in the CoT; and (2) P and A do not introduce any new substantive statements absent from the CoT. This ensures that, semantically, the CoT entails both P and A. We also checked whether the final answer in A is equivalent to the final answer derived in CoT.

For OpenThoughts, we found that in $99\%$ and $89\%$ of the 100 sampled examples, the CoT semantically entails P and A, respectively. For Bespoke, these figures are $97\%$ and $93\%$. The few exceptions where CoT does not entail A are due to new information introduced in A. They primarily involve open-ended science or coding questions where the answer is not easily stored in a latex box\footnote{Example: What is the effect of different types of amino acid residues in the binding site of protein on the stability of DNA-protein complex in molecular dynamics simulations?}. However, in $100\%$ and $99\%$ of examples in OpenThoughts and Bespoke respectively, the final answer in A is equivalent to the final answer in the CoT section. That indicates that despite these examples having new information in answer section, their final answer still aligns with that in CoT section. This explains why adding supervision on P and A provides minimal benefit over CoT alone. For details of the design and results of linguistic analysis, refer to Appendix \ref{appendix:linguistic_analysis}.

\begin{figure*}[!t]
  \includegraphics[width=1.0\linewidth]{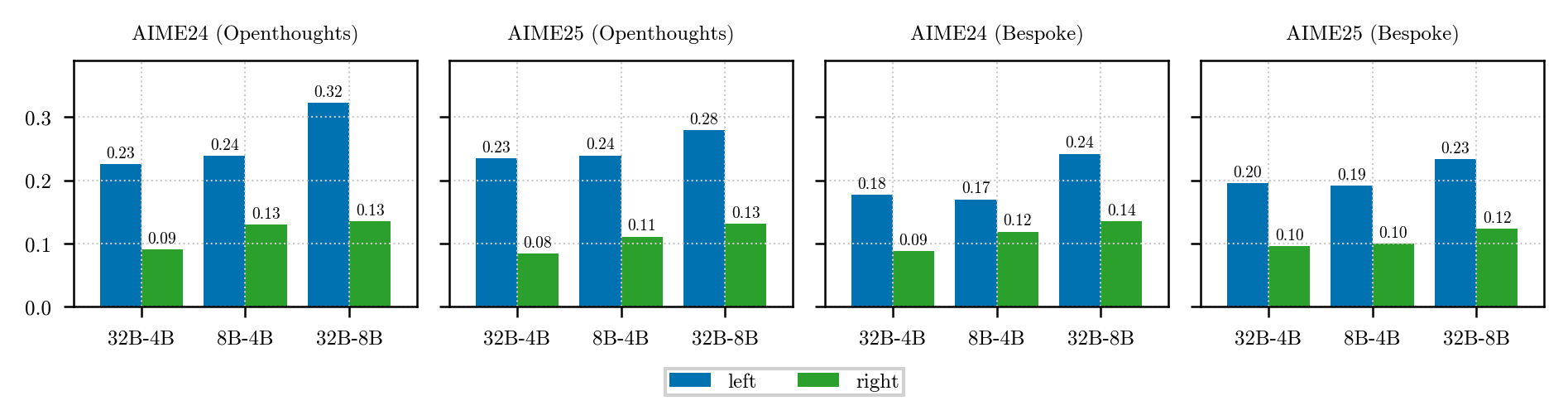}
  \caption {Impact of $50\%$ budget allocation strategies (left, right) on model accuracy. Retaining the left portion of the sequence consistently outperforms the right, indicating that early tokens are more valuable for learning than the later tokens. This ablation of training budget verifies that the efficiency of using 50\% tokens is not a general
trait when reducing the number of training tokens.}\label{fig:combined_half_keep_strategy}
\end{figure*}

\begin{figure}[!t]
  \includegraphics[width=1.0\linewidth]{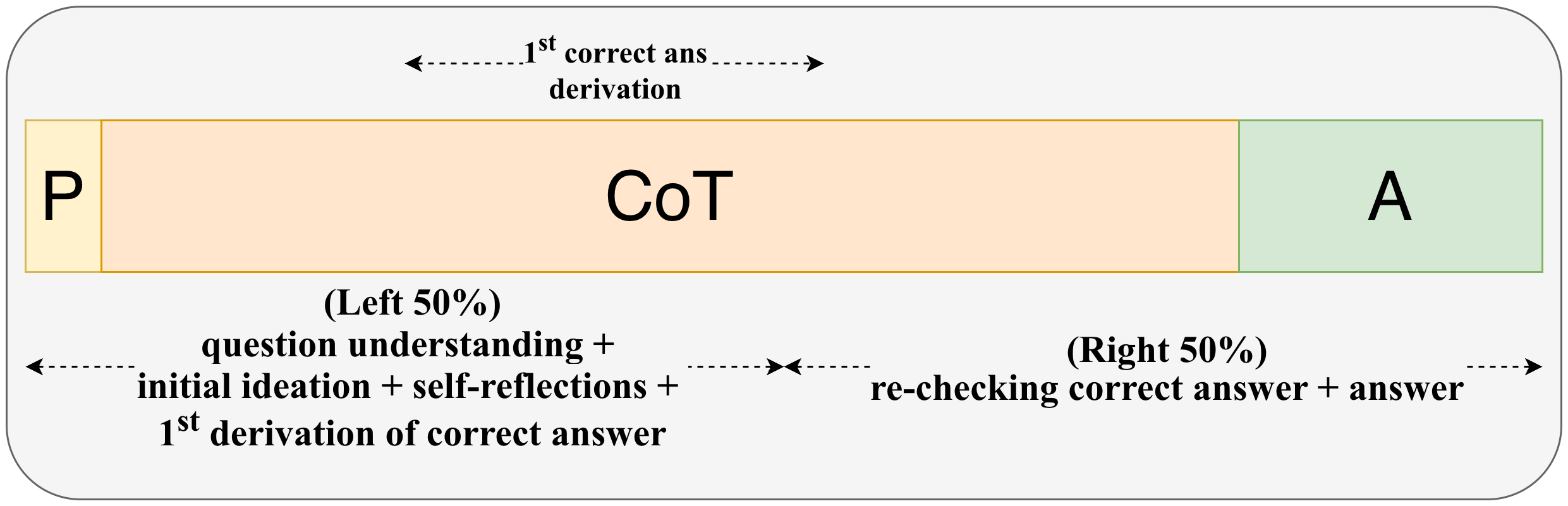}
  \caption {First half of tokens tends to contain denser useful information than the second half, in line with stronger performance when training on the left $50\%$ of tokens. Complete examples, each for one of the five datasets, are included in Appendix~\ref{appendix:full_training_examples}.}\label{fig:cot_first_correct_ans_derivation}
\end{figure}

\subsection{Early vs. Late Tokens (RQ2)}\label{subsec:scaling_behavior}

\paragraph{Marginal benefits of longer reasoning sequences.} Figure~\ref{fig:combined_LSP} shows that across all distillation settings, LSP exhibits similar scaling behavior of the student model performance. They show nearly monotonic accuracy improvement as LSP increases and then saturates. The flattening near the right edge suggests those trimmed tokens are not required for effective distillation, implying substantial compute savings with minimal performance drop.

\paragraph{Role of model size.} The 32B-8B setting is consistently the top performer, and the ranking is largely determined by student size: 8B > 4B > 1.7B > 0.6B. For a fixed student size, curves from different teachers track closely. For the 4B student model, it performs comparably and sometimes better when distilling from 8B rather than 32B model, which echoes the findings of \citet{li-etal-2025-small-models} that smaller models do not necessarily learn better from much larger teachers. For small students (0.6B and 1.7B), training with LSP does not reliably beat the baselines. Their curves hew close to the dotted baselines across $p$, suggesting limited capacity to absorb additional supervision. The benefits of LSP mainly materialize once the student has enough scale (4B and 8B). Therefore, in later experiments, these two small students will not be involved.

\begin{table}[h]
\centering
\resizebox{\columnwidth}{!}{%
\begin{tabular}{@{}l r r r r @{}}
\toprule
& \multicolumn{2}{c}{Openthoughts} & \multicolumn{2}{c}{Bespoke}\\
\cmidrule(lr){2-3}\cmidrule(lr){4-5}
Pairs & AIME24 & AIME25 & AIME24 & AIME25 \\
\midrule
32B--4B & 0.5 & 0.5 & 0.6 & 0.2 \\
8B--4B  & 0.4 & 0.2 & 0.2 & 0.2 \\
32B--8B & 0.6 & 0.5 & 0.4 & 0.5 \\
\bottomrule
\end{tabular}%
}
\caption{Knees for different experimental setups to indicate the most cost-effective LSP.}
\label{tab:openthoughts_knee}
\end{table}

\textbf{Deciding the most cost-effective budget.} To anchor the token budget allocation, we select $p=0.5$ as our target. This decision is driven by three factors: performance saturation, computational efficiency, and experimental rigor. First, we identified the most cost-effective point (``knee'') in the accuracy-vs-LSP curves with ~\citet{satopaa2011finding}\footnote{\url{https://github.com/arvkevi/kneed}}, as shown in Table~\ref{tab:openthoughts_knee}. We observe that across all settings, the knee consistently occurs at $p \le 0.6$, with the majority at or below $p=0.5$.
Second, as detailed in Section \ref{sec:compute_efficiency}, $p=0.5$ serves as a critical boundary before computational costs begin to rise drastically.
Third, selecting $p=0.5$ enables a clean, symmetrical ablation between the Left 50\% and Right 50\% of the sequence. A budget of $p > 0.5$ would necessitate overlapping tokens. As seen in Table~\ref{tab:0p5_1p0_ratio_openthoughts}, across all 30 settings (3 teacher-student pairs, 5 datasets, and 2 benchmarks), training with the first half of tokens retains on average $\approx91\%$ of the accuracy when compared to training with all tokens, with retention being as high as $\approx99\%$ for AIME25 with 8B-4B setup.

\begin{table}[ht]
\centering
\resizebox{\columnwidth}{!}{%
\begin{tabular}{@{}l r r r r r r r r r r@{}}
\toprule
& \multicolumn{2}{c}{OpenThoughts} & \multicolumn{2}{c}{Bespoke} & \multicolumn{2}{c}{Synthetic-1}&\multicolumn{2}{c}{Nemotron}&\multicolumn{2}{c}{SkyT1}\\
\cmidrule(lr){2-3}\cmidrule(lr){4-5}\cmidrule(lr){6-7}\cmidrule(lr){8-9}\cmidrule(lr){10-11}
Pairs & \multicolumn{1}{c}{24} & \multicolumn{1}{c}{25} & \multicolumn{1}{c}{24} & \multicolumn{1}{c}{25} & \multicolumn{1}{c}{24} & \multicolumn{1}{c}{25} & \multicolumn{1}{c}{24} & \multicolumn{1}{c}{25} & \multicolumn{1}{c}{24} & \multicolumn{1}{c}{25}\\
\midrule

32B-4B & 96.21 & 97.40 & 97.98 & 93.28 & 79.28 & 91.77 & 96.56 & 90.71 & 87.41 & 89.65 \\
8B-4B & 98.07 & 98.92 & 85.79 & 87.62 & 82.54 & 92.12 & 91.50 & 91.62 & 75.23 & 89.01 \\
32B-8B & 95.96 & 97.63 & 92.43 & 87.16 & 95.23 & 97.06 & 92.21 & 96.63 & 80.12 & 82.66 \\

\bottomrule
\end{tabular}%
}
\caption{The performance retention rate (\%) training with the first half of the tokens of each example in five datasets, compared to training with full-length examples. Performance is evaluated on AIME24 (``24'') and AIME25 (``25''). See Appendix \ref{appendix:raw_exp_res_0p5_1p0_five_datasets} for the raw values used to compute the retention rate.} 
\label{tab:0p5_1p0_ratio_openthoughts}
\end{table}

\begin{figure*}[!t]
  \includegraphics[width=1.0\linewidth]{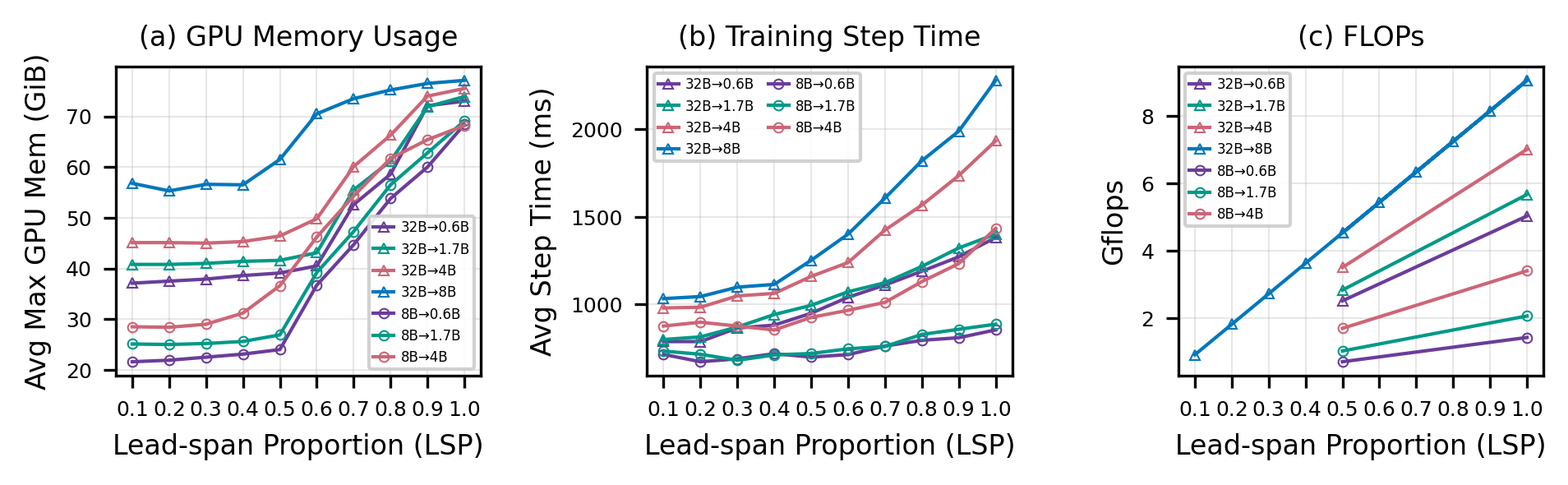}
  \caption {GPU memory usage, time spent for each training step, and FLOPs as functions of LSP. Triangles indicate a 32B teacher, circles indicate an 8B teacher, and matching colors indicate the same student model. Due to the linear behavior of flops, not all but only selected LSPs ($0.5, 1.0)$ for some teacher-student pairs are profiled.}\label{fig:combined_LSP_flops_mem_stepTime}
  \vspace{-2mm}
\end{figure*}

\textbf{Why Early Tokens Suffice?} The fact that the first $50\%$ of tokens yields similar performance compared to full-length raises a mechanistic question: does the early segment already contain the key reasoning signal? To probe this, we analyze where the \emph{first derivation of the correct answer} appears inside the CoT, treating it as a proxy for the earliest point at which the reasoning becomes sufficient. For each of $100$ randomly sampled training examples, we prompt GPT-5.1 to identify the first derivation of the correct answer, and then compute its relative position within the sequence.

Across datasets, the first correct derivation tends to occur near the midpoint of the whole sequence; see Appendix~\ref{appendix:linguistic_analysis} for details. We illustrate this phenomenon in Figure~\ref{fig:cot_first_correct_ans_derivation}. This near-midpoint tendency aligns with the accuracy--vs--LSP curves exhibiting knees at $p \le 0.6$ (Table~\ref{tab:openthoughts_knee}) and with the strong performance retention at $p=0.5$ (Table~\ref{tab:0p5_1p0_ratio_openthoughts}). Together, these results suggest that an early budget captures most of the learnable signal for distillation.

Qualitative analysis of the content preceding the first derivation shows that this segment contains question understanding, initial ideation, several rounds of self-reflections, and substantial progress toward the final answer. In many examples, the remaining tokens are then largely dominated by verification, minor algebraic clean-up, and re-statements of an already-correct answer, which appear less informative for distillation than the earlier reasoning steps.

We operationalize self-reflection, a key contributor to the capabilities of reasoning models \cite{muennighoff2025s1, li2025llmseasilylearnreason}, by counting occurrences of lexical cues (e.g., ``wait'', and related variants; see full list in Appendix~\ref{appendix:linguistic_analysis_methodology}) that appear before the first correct derivation point. On average, we observe $2.2$ (OpenThoughts), $2.4$ (Bespoke), $2.8$ (Synthetic-1), $5.1$ (Nemotron), and $1.9$ (SkyT1) self-reflection cues before the first correct derivation.

\textbf{Ablation Studies}
Having established that the left $50\%$ often encapsulates the core reasoning process, we use the Left vs. Right ablation to verify that the efficiency gains are specific to this self-contained initial segment, rather than a general property of reducing the token count. We compare training with the Left $50\%$ against the Right $50\%$. As shown in Figure~\ref{fig:combined_half_keep_strategy}, training with the left $50\%$ of the sequence consistently and significantly outperforms training with the right $50\%$. This is further supported by our loss analysis (Appendix~\ref{appendix:loss_analysis}), which reveals that the loss on the left 50\% is comparable to the full sequence, whereas the loss on the right 50\% is relatively higher, indicating that the initial segment is more amenable to learning.

Furthermore, we investigate the sensitivity of our method to the distillation weight $\lambda$. Our experimental results show that the approach is robust, as performance remains consistently strong across $\lambda \in \{0.25, 0.5, 0.75\}$. This consistency demonstrates that the effectiveness stems from the proposed method itself, rather than relying on the specific $\lambda=0.5$ selection. For detailed ablation results, see Appendix~\ref{appendix:ablation_studies}.

\subsection{Compute Efficiency}\label{sec:compute_efficiency}
Computationally, our $50\%$ training protocol yields substantial resource savings, as illustrated in Figure~\ref{fig:combined_LSP_flops_mem_stepTime}. We observe a critical inflection point at LSP $=0.5$ for both GPU memory usage\footnote{GPU memory is measured with the NVIDIA Data Center GPU Manager (DCGM) metric DCGM\_FI\_DEV\_FB\_USED to reflect practical device-level VRAM usage.} and training step time, beyond which costs escalate rapidly due to the quadratic complexity of attention mechanisms.

\textbf{Memory Efficiency.} Peak GPU memory usage remains relatively modest for LSP $\leq 0.5$ but spikes sharply thereafter. For instance, in the 8B$\to$4B setting, processing the full sequence ($p=1.0$) consumes up to 68.2 GiB, necessitating an 80GB H100 GPU. By truncating to $50\%$, peak memory drops to 36.6 GiB, fitting comfortably on a standard 40GB A100 GPU. This reduction lowers the hardware barrier significantly, enabling distillation on more accessible infrastructure.

\textbf{Training Speed.} Training step time follows a similar trajectory. While larger students generally require more time, we also observe that teacher size imposes a distinct overhead; for the same 4B student, distilling from a 32B teacher is consistently slower than from an 8B teacher. By capping the sequence at $50\%$, we achieve a roughly $2\times$ reduction in both training step time and FLOPs, confirming that $p=0.5$ is a practical sweet spot for resource-constrained distillation.

\section{Conclusion}
In this work, we demonstrate that the reasoning trace itself is the dominant carrier of supervisory signal in knowledge distillation. By systematically varying the supervised sections of training examples, we found that models trained on the reasoning trace alone achieve performance comparable to full-sequence supervision. Moreover, our investigation into token budget allocation reveals that critical reasoning behaviors are concentrated in early tokens; truncating examples to just their first $50\%$ of tokens retains $\approx 91\%$ of downstream accuracy on math tasks while reducing training time and memory usage by $\approx 50\%$. This establishes sequence truncation as a powerful efficiency lever, positioning partial-sequence training as a cost-effective strategy for scaling reasoning distillation.

\section*{Limitations}
Our study focuses on mathematical reasoning tasks. While mathematics provides a rigorous testbed for reasoning with verifiable ground truth, extending these findings to other reasoning domains (e.g., commonsense reasoning, and code generation) remains an avenue for future work. Our conclusions are drawn from highly structured datasets, and may not directly generalize to less structured or interactive reasoning settings.

We conduct experiments within the Qwen3 model family, using Qwen3-32B and Qwen3-8B as teachers and Qwen3-8B/4B/1.7B/0.6B as students. This controlled setup isolates the effects of sequence truncation from confounding factors introduced by architectural heterogeneity across model families. Investigating cross-family distillation (e.g., Qwen to Gemma) represents a natural extension.

In our LSP comparisons, we train on the same set of examples across all settings, and only change how much of each example is kept. This means the total number of training tokens is not matched across LSP values: for instance, LSP $=0.5$ uses about half as many training tokens as LSP $=1.0$. Therefore, our study does not test an equal-token-budget scenario (e.g., using shorter sequences but proportionally more examples). Evaluating this fixed-budget tradeoff is an important direction for future work.

\section*{Ethical Considerations}
This work investigates efficient training protocols for knowledge distillation. All datasets and models used are publicly available and were created by prior work for research purposes.

Our findings on sequence truncation have potential positive implications for democratizing access to reasoning model training, as they substantially reduce computational requirements. By demonstrating that comparable performance can be achieved with approximately half the training compute and memory, this work may lower barriers for researchers and practitioners with limited computational resources.

We acknowledge that improvements in reasoning model distillation could, in principle, be applied to both beneficial and harmful use cases. However, the efficiency techniques we propose do not fundamentally alter model capabilities. They reduce training costs for existing distillation paradigms. We believe the benefits of making reasoning distillation more accessible outweigh potential risks.

\section*{Acknowledgments}
Muhammad Abdul-Mageed acknowledges support from Canada Research Chairs (CRC), the Natural Sciences and Engineering Research Council of Canada (NSERC; RGPIN-2018-04267), the Social Sciences and Humanities Research Council of Canada (SSHRC; 895-2020-1004; 895-2021-1008), Canadian Foundation for Innovation (CFI; 37771), Digital Research Alliance of Canada\footnote{\href{https://alliancecan.ca}{https://alliancecan.ca}}, UBC Advanced Research Computing-Sockeye\footnote{\href{https://arc.ubc.ca/ubc-arc-sockeye}{https://arc.ubc.ca/ubc-arc-sockeye}}, and UBC Institute for Computing, Information and Cognitive Systems (ICICS)\footnote{\href{https://icics.ubc.ca/}{https://icics.ubc.ca/}}.

\bibliography{custom}

\newpage
\onecolumn
\appendix
\section{Appendix}\label{sec:appendix}

\subsection{Training Dataset Information }\label{appendix:train_dataset_info}
\subsubsection{Full Training Examples}\label{appendix:full_training_examples}

\begin{figure}[!ht]
  \centering
  \includegraphics[width=0.95\linewidth]{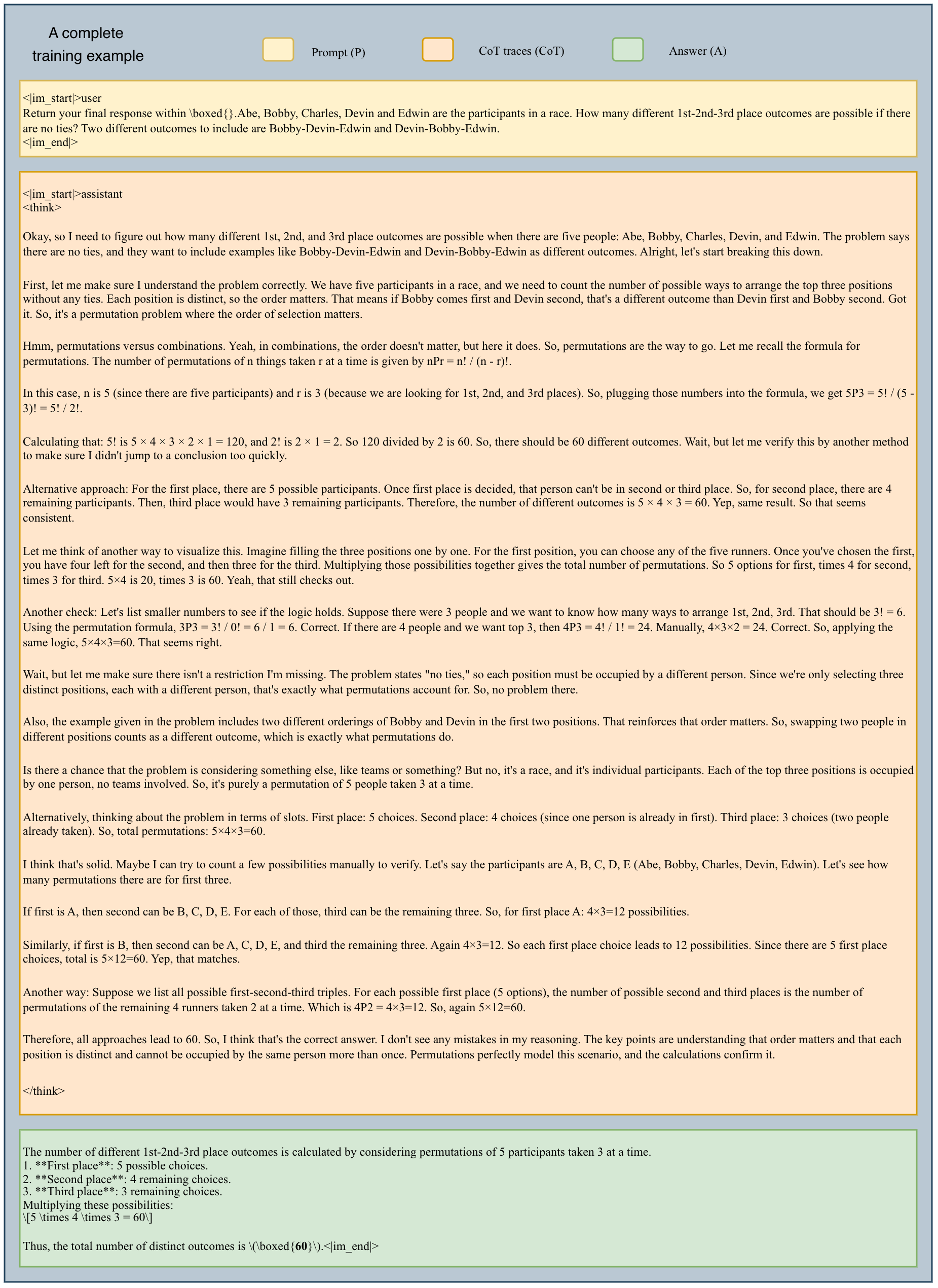}
  \caption {A complete training example from Bespoke with three sections divided for illustration purposes.}\label{fig:complete_training_ex_Abe}
\end{figure}

\begin{figure}[!ht]
  \includegraphics[width=1.0\linewidth]{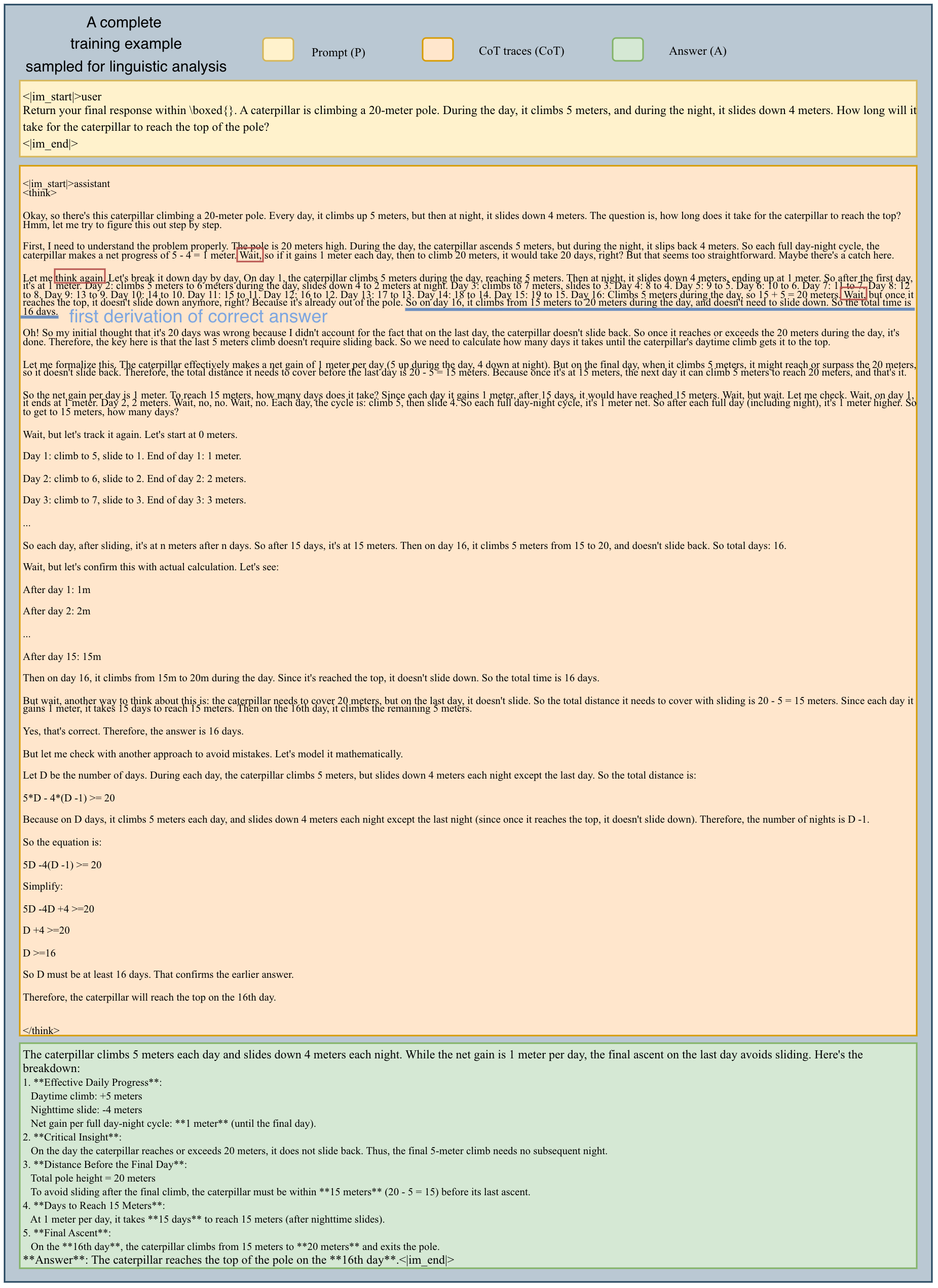}
  \caption {A complete training example from Openthoughts sampled for linguistic analysis with three sections divided for illustration purposes. Red boxes indicate self-reflection keywords. Blue underline is the first derivation of correct answer labeled by GPT-5.1.}\label{fig:complete_training_ex_caterpillar}
\end{figure}

\begin{figure}[!ht]
  \includegraphics[width=1.0\linewidth]{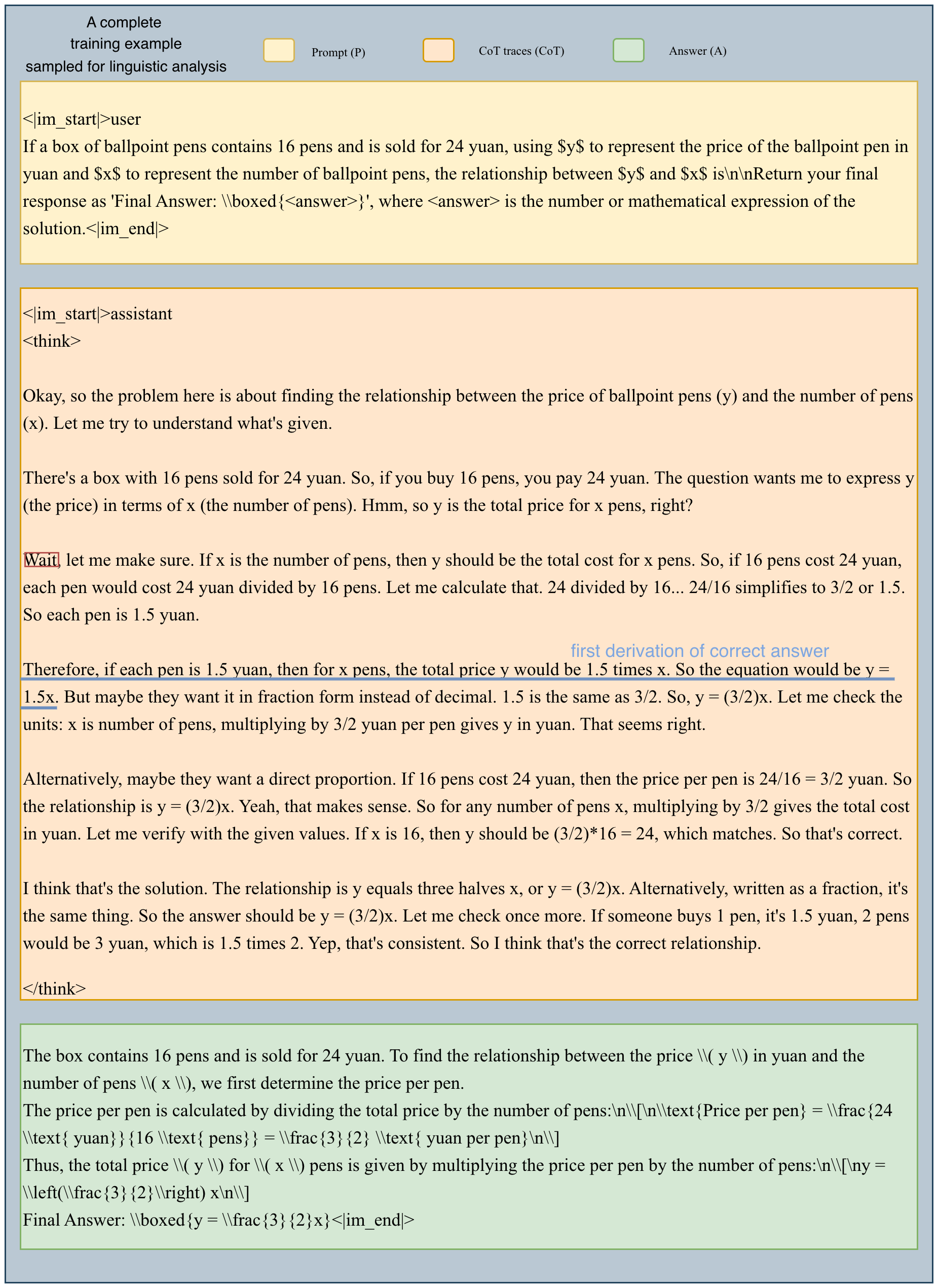}
  \caption {A complete training example from Synthetic-1 sampled for linguistic analysis with three sections divided for illustration purposes. Red boxes indicate self-reflection keywords. Blue underline is the first derivation of correct answer labeled by GPT-5.1.}\label{fig:complete_training_ex_synthetic1_pen}
\end{figure}

\begin{figure}[!ht]
  \includegraphics[width=1.0\linewidth]{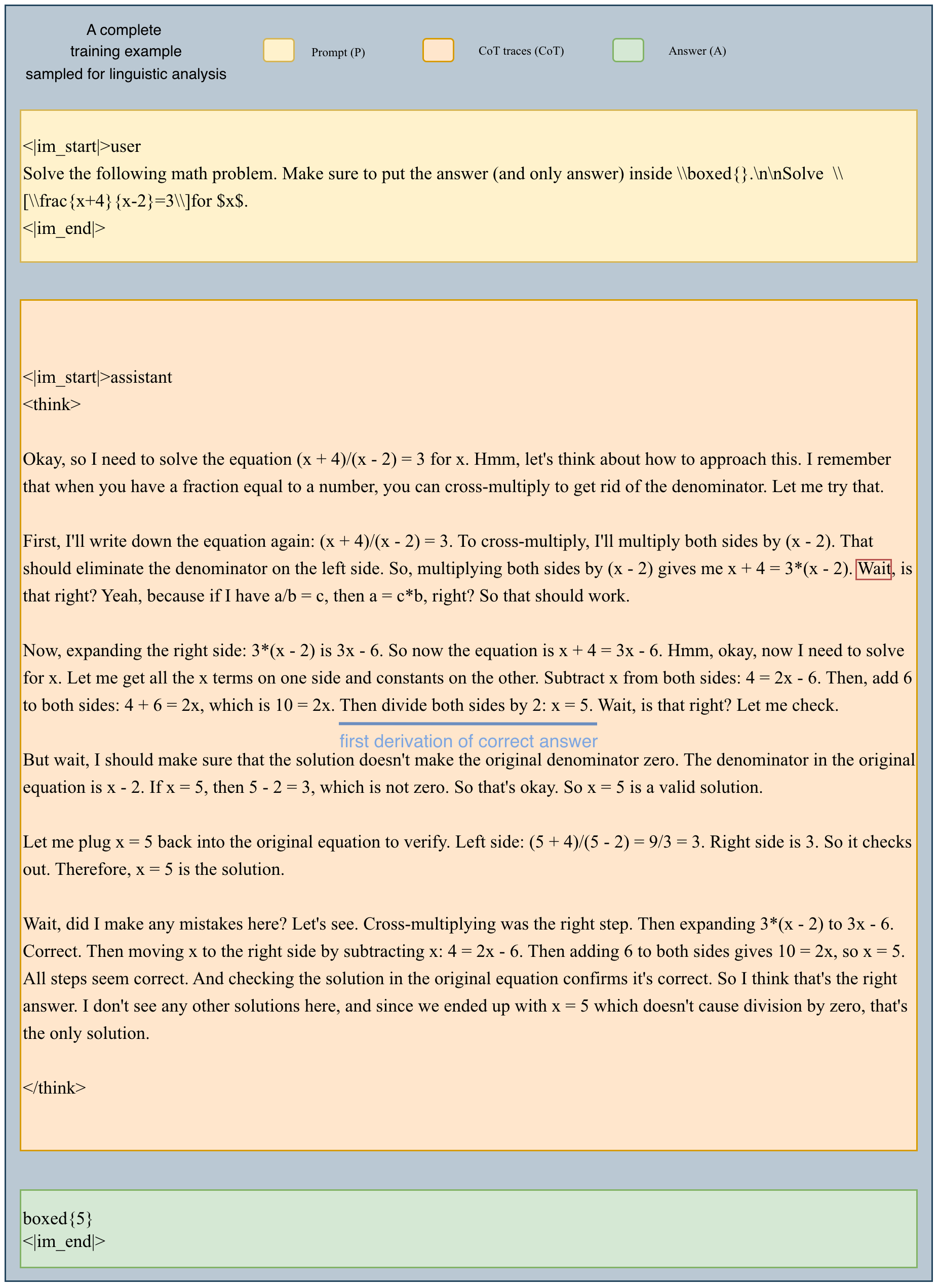}
  \caption {A complete training example from Nemotron  sampled for linguistic analysis with three sections divided for illustration purposes. Red boxes indicate self-reflection keywords. Blue underline is the first derivation of correct answer labeled by GPT-5.1.}\label{fig:complete_training_ex_nemotron_5}
\end{figure}

\begin{figure}[!ht]
  \includegraphics[width=1.0\linewidth]{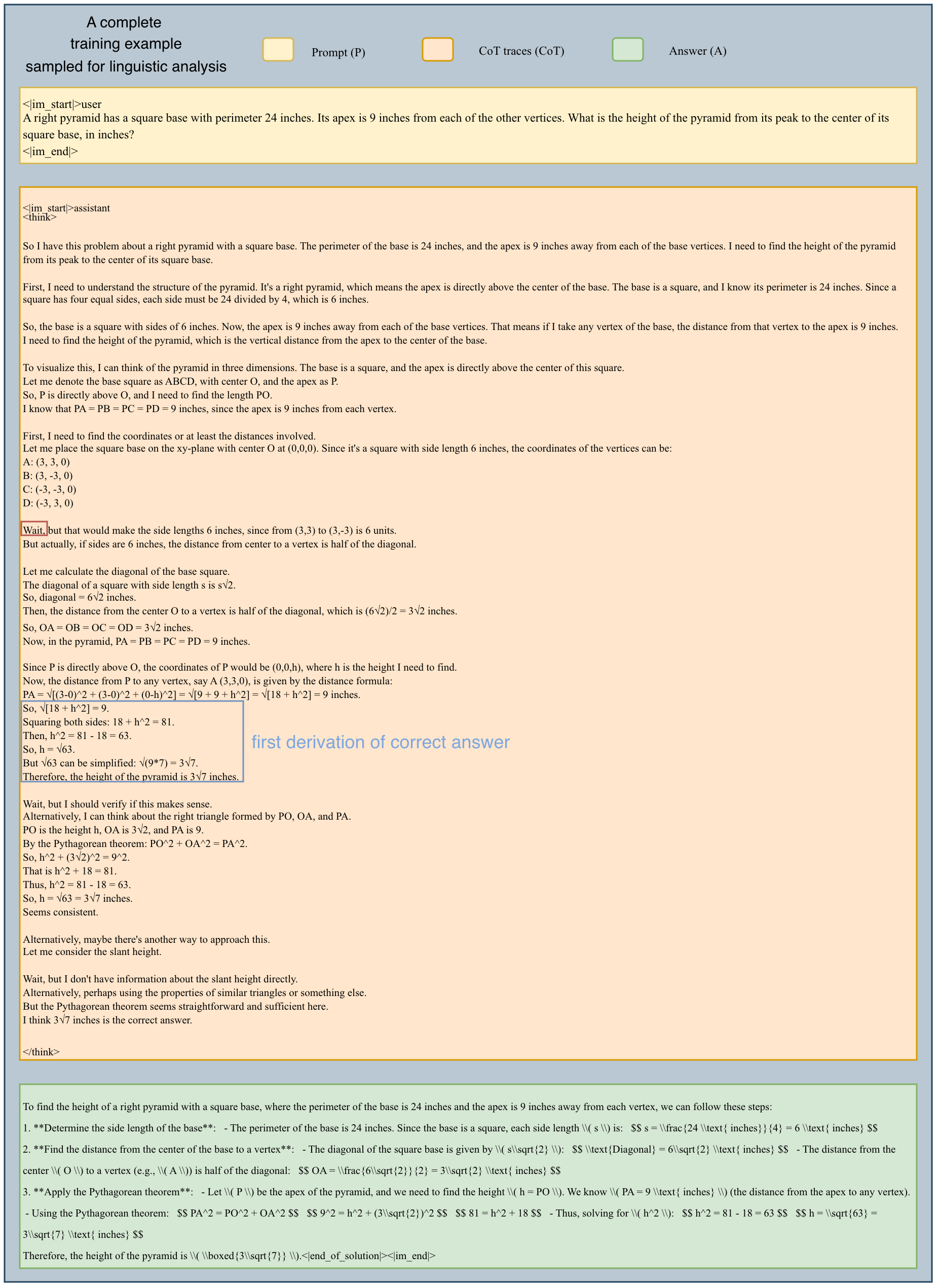}
  \caption {A complete training example from SkyT1 sampled for linguistic analysis with three sections divided for illustration purposes. Red boxes indicate self-reflection keywords. Blue underline is the first derivation of correct answer labeled by GPT-5.1.}\label{fig:complete_training_ex_pyramid}
\end{figure}

\subsubsection{Training Dataset Token Count Distribution}\label{appendix:train_set_token_count_dist}

\begin{figure}[!ht]
    \centering
  \includegraphics[width=0.9\linewidth]{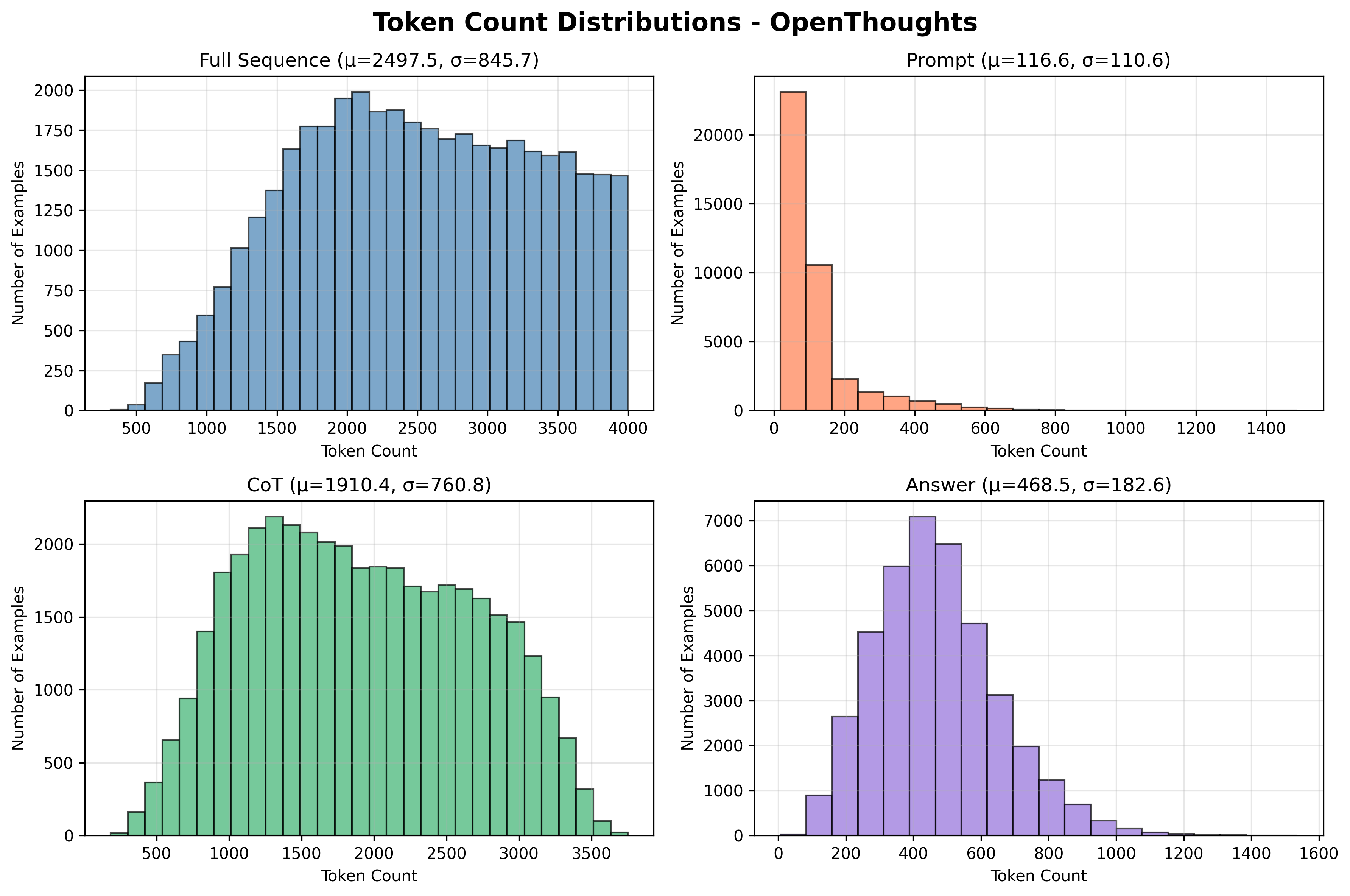}
  \caption {Token count distributions for prompts, chain-of-thought, answers, and full sequences in the sampled 40k training examples from OpenThoughts-114k dataset.}\label{fig:token_dist_openthoughts}
\end{figure}

\begin{figure}[!ht]
    \centering
  \includegraphics[width=0.9\linewidth]{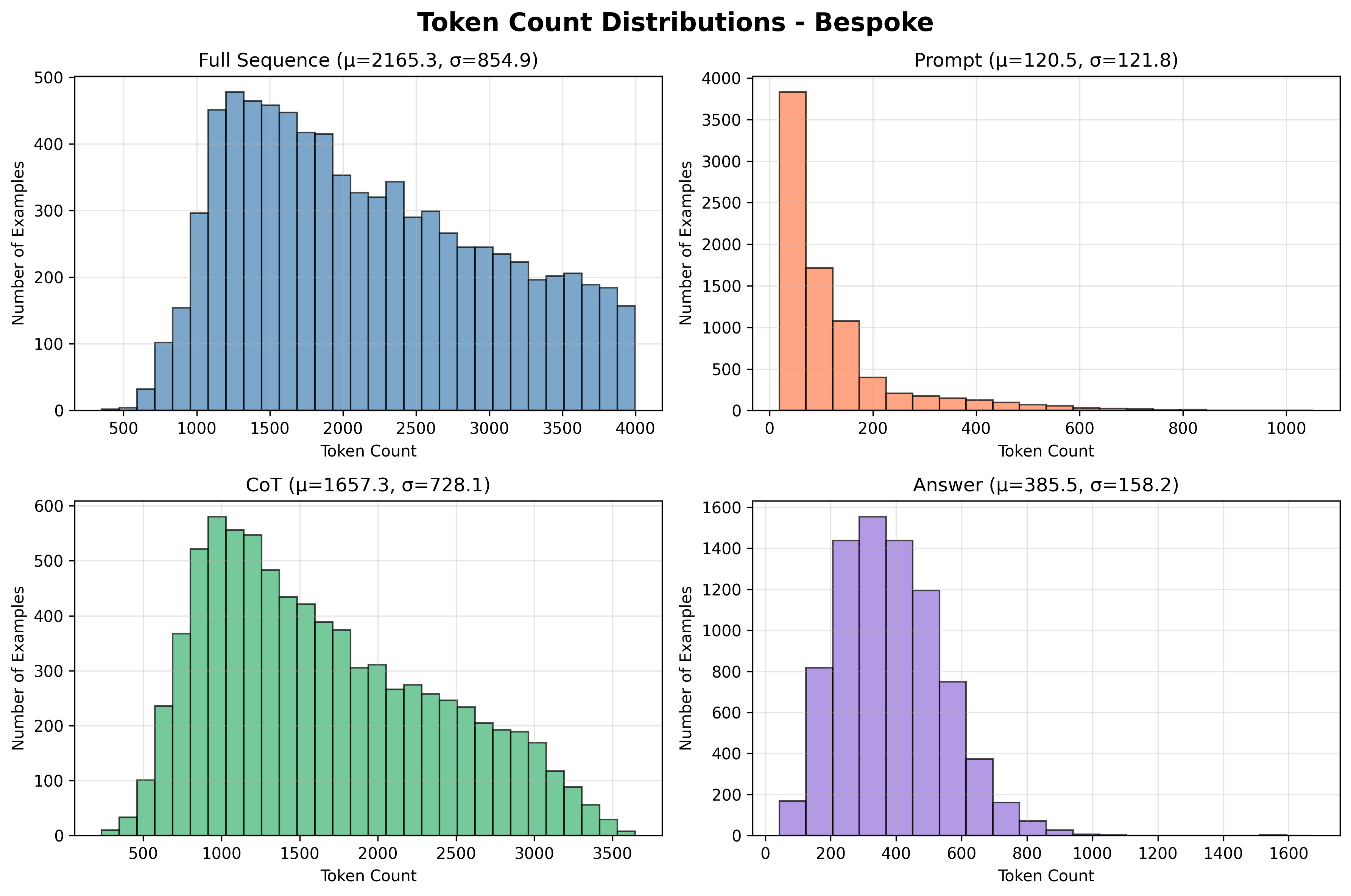}
  \caption {Token count distributions for prompts, chain-of-thought, answers, and full sequences in the sampled 8k training examples from Bespoke-Stratos-17k dataset.}\label{fig:token_dist_bespoke}
\end{figure}

\begin{figure}[!ht]
    \centering
  \includegraphics[width=0.9\linewidth]{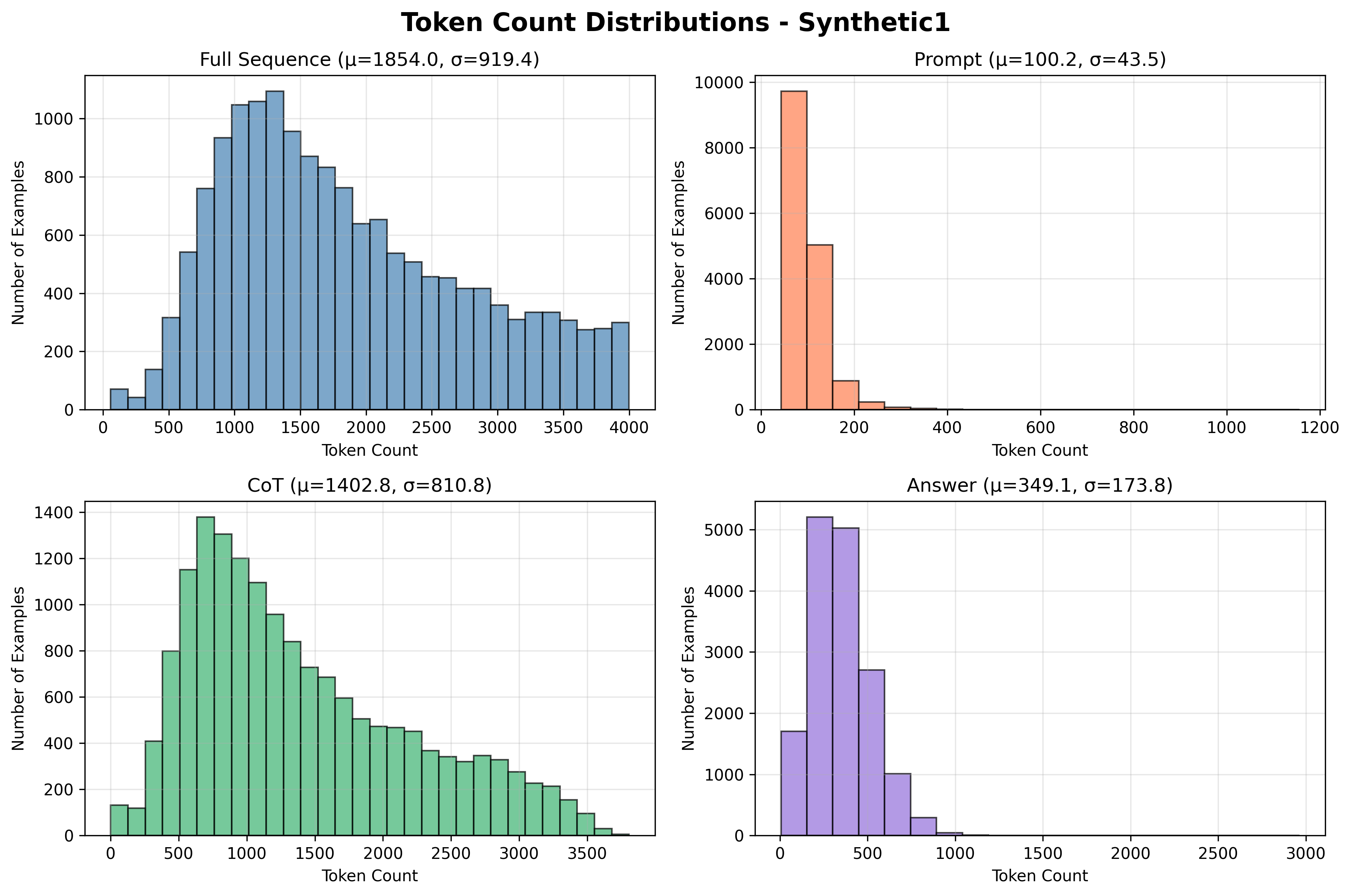}
  \caption {Token count distributions for prompts, chain-of-thought, answers, and full sequences in the sampled 16k training examples from Synthetic-1  dataset.}\label{fig:token_dist_synthetic1}
\end{figure}

\begin{figure}[!ht]
    \centering
  \includegraphics[width=0.9\linewidth]{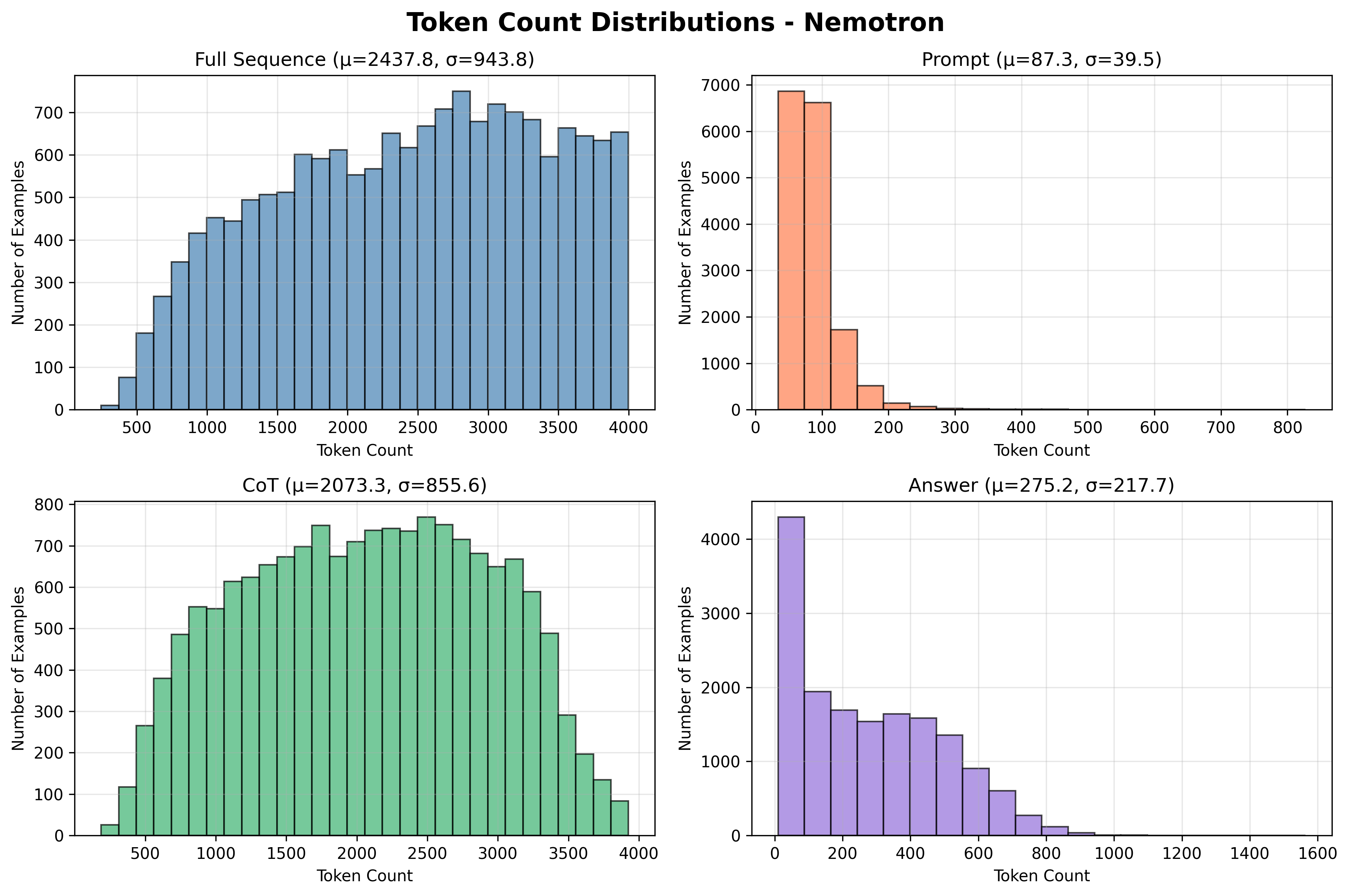}
  \caption {Token count distributions for prompts, chain-of-thought, answers, and full sequences in the sampled 16k training examples from Llama-Nemotron-Post-Training-Dataset   dataset.}\label{fig:token_dist_nemotron}
\end{figure}

\begin{figure}[!ht]
    \centering
  \includegraphics[width=0.9\linewidth]{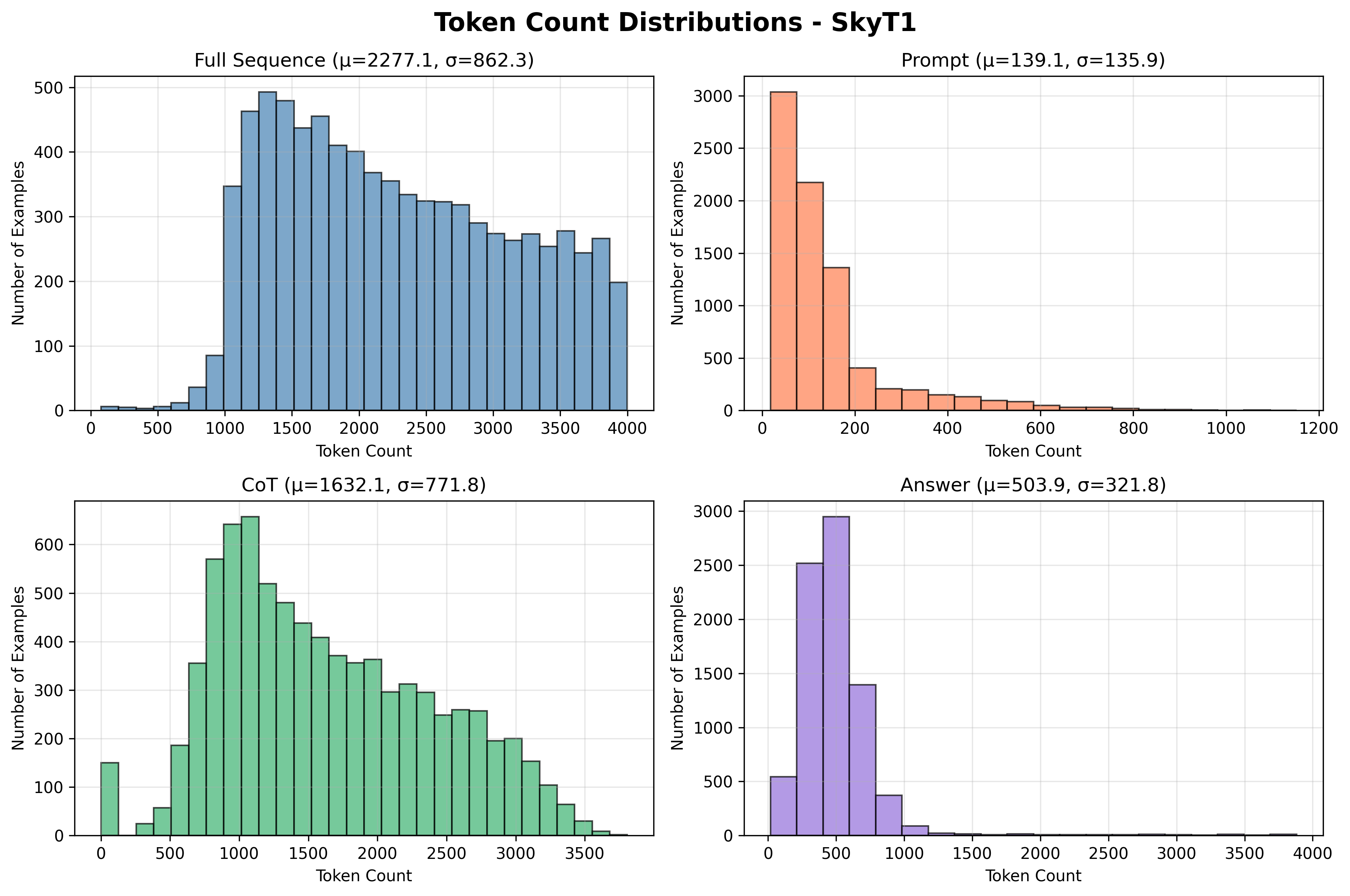}
  \caption {Token count distributions for prompts, chain-of-thought, answers, and full sequences in the sampled 8k training examples from SkyT1-17k dataset.}\label{fig:token_dist_skyt1}
\end{figure}

\begin{table}[ht]
    \footnotesize
    \centering
    \begin{tabularx}{\textwidth}{|>{\ttfamily\raggedright\arraybackslash}X|}
        \hline
         You are given a question and two texts, t1 and t2.\newline
\newline
Question:\newline
--------------------\newline
\{question\_text\}\newline
--------------------\newline
\newline
Text t1:\newline
--------------------\newline
\{t1\_text\}\newline
--------------------\newline
\newline
Text t2:\newline
--------------------\newline
\{t2\_text\}\newline
--------------------\newline
\newline
Your tasks:\newline
\newline
1. Check whether the Question is fully covered by t1.\newline
2. Check whether t2 is fully covered by t1.\newline
3. Check whether the final boxed answer in t2 is also the final answer in t1 (i.e., the answer that t1 itself treats as its final conclusion).\newline
4. Find the earliest place in t1 where the final answer is actually derived/computed from the reasoning. Return a small continuous substring (one sentence or short phrase) that contains that derivation.\newline
\newline
Definitions:\newline
\newline
- "Text A is fully covered by Text B" means that every substantive statement in A can be found in B (possibly with different wording or notation), and A does not introduce any new substantive statements absent from B.\newline
\newline
When comparing content:\newline
\newline
- Treat synonymous wording and mathematically equivalent expressions as the same (e.g., 0.5, 1/2, and "one half" are equivalent).\newline
- Ignore differences in order, formatting, and minor paraphrasing that do not change meaning.\newline
\newline
Final answer logic:\newline
\newline
Final answer in a piece of text:\newline
- Look for LaTeX-style boxes: \\boxed\{\{...\}\} in the text or the last explicit conclusion that clearly answers the question.\newline
\newline
Determine whether t2's final answer is equivalent to t1's final answer, up to differences in notation or wording. This must reflect that t1 itself presents that value/expression as its final conclusion, not merely as an intermediate value.\newline
\newline
Derivation/computation of final answer (for Task 4):\newline
\newline
1. Final answer value: Extract from t2 (look for \\boxed\{\{...\}\} or the explicit final answer statement).\newline
\newline
2. "Derived/computed": The earliest step in t1 where the final answer is derived or computed with math or logic (e.g. "3! = 6", "so there are 6 outcomes", "thus the answer is 6" after reasoning).\newline
\newline
3. What to ignore:\newline
   - Later repetitions like "Therefore, the answer is 6" if an earlier sentence already computed 6.\newline
   - Hypotheticals or guesses (e.g. "if the answer were 6 ..." / "maybe it's 6").\newline
   - Mentions of the answer taken from the problem statement, not from the reasoning.\newline
\newline
4. Substring rules:\newline
   - Must be one contiguous block of text from t1.\newline
   - Must include the first valid derivation of the final answer.\newline
   - Copy it **verbatim**: do not rephrase or correct; preserve spacing, punctuation, line breaks.\newline
\newline
Output format:\newline
\newline
Return your result only as a JSON object with exactly four keys:\newline
\newline
- is\_question\_fully\_covered\_by\_t1 (boolean):\newline
  - true if the Question is fully covered by t1.\newline
  - false otherwise.\newline
\newline
- is\_t2\_fully\_covered\_by\_t1 (boolean):\newline
  - true if t2 is fully covered by t1.\newline
  - false otherwise.\newline
\newline
- is\_t2\_final\_answer\_considered\_final\_in\_t1 (boolean):\newline
  - true if the content of t2's final answer is equivalent to t1's final answer as defined above.\newline
  - false otherwise.\newline
\newline
- first\_derivation (string):\newline
  - The verbatim contiguous substring from t1 that contains the first derivation of the final answer.\newline
  - Empty string ("") if such derivation cannot be found in t1.\newline
\newline
Do not include any other explanations, reasoning, or extra fields.\newline
Respond with only valid JSON, for example:\newline
\newline
\{\{\newline
  "is\_question\_fully\_covered\_by\_t1": true,\newline
  "is\_t2\_fully\_covered\_by\_t1": true,\newline
  "is\_t2\_final\_answer\_considered\_final\_in\_t1": false,\newline
  "first\_derivation": "Therefore, 3! = 3 x 2 x 1 = 6."\newline
\}\}
         \\ \hline
    \end{tabularx}
    \caption{Instruction to conduct linguistic analysis with GPT-5.1}\label{tab:linguistic_analysis_instruction}
\end{table}

\clearpage
\subsection{Linguistic Analysis}\label{appendix:linguistic_analysis}
\subsubsection{Methodology}\label{appendix:linguistic_analysis_methodology}
To mechanistically understand why supervising the Prompt (P) and Answer (A) sections yields minimal gain over supervising the Chain-of-Thought (CoT) alone (RQ1), and why the first $50\%$ of tokens are sufficient for effective distillation (RQ2), we designed an automated linguistic analysis pipeline. We randomly sampled $100$ training examples from the training split of each dataset. Each example was analyzed using \texttt{gpt-5.1-2025-11-13} as a judge to verify semantic entailment and to locate critical reasoning steps. The complete instruction is shown in Table~\ref{tab:linguistic_analysis_instruction}. We use generic terms ``t1'' and ``t2'', instead of explicitly using ``Chain-of-thought'' and ``Answer'', to refer to CoT section and answer section to avoid biased judgments.

\paragraph{Semantic Entailment Analysis.}
To determine if the information in P and A is redundant given the CoT, we prompted the judge model to evaluate three conditions:
\begin{enumerate}
    \item \textbf{Prompt Coverage:} Whether every substantive statement in the Prompt is fully covered in the CoT.
    \item \textbf{Answer Coverage:} Whether every substantive statement in the Answer is fully covered in the CoT.
    \item \textbf{Final Answer Equivalence:} Whether the final boxed answer in the Answer section is equivalent to the final conclusion derived in the CoT.
\end{enumerate}
We defined ``fully covered'' to mean that all substantive information in the target text (P or A) is found in the source text (CoT), allowing for rephrasing but disallowing new information.

\paragraph{Locating the First Derivation of Correct Answer.}
To investigate the concentration of reasoning signals (RQ2), we asked the judge to identify the \emph{earliest} point in the CoT where the correct final answer is  derived. The model was instructed to extract a verbatim contiguous substring corresponding to this derivation step (e.g., ``Therefore, $3! = 6$'').

To map this substring to a relative position within the sequence, given this substring, we programmatically locate its character position within the CoT, tokenize the full CoT using the Qwen3 tokenizer, and map the character position to its corresponding token index. The relative position is then computed as the ratio of this token index to the total number of tokens in the CoT, yielding a value between 0 and 1 that indicates where in the CoT traces the correct final answer is first derived.

\paragraph{Self-Reflection Analysis.}
To quantify the presence of self-correction prior to the final solution, we counted the occurrences of specific self-reflection keywords in the CoT text preceding the character position of the identified first derivation of correct answer. The set of keywords used is collected from \citet{liu2025understandingr1zeroliketrainingcritical, baek2025understandingdistilledreasoningmodels}:
\begin{itemize}
    \item \textit{recheck, rethink, reassess, reevaluate, re-evaluate, reevaluation, re-examine, reexamine, reconsider, reanalyze, double-check, check again, think again, verify again, go over the steps, wait}.
\end{itemize}
This metric serves as a proxy for the density of critical reasoning and error-correction behaviors in the earlier parts of the chain of thought.

\subsubsection{Results}
We present the results of our linguistic analysis on all five datasets in Table~\ref{tab:linguistic_analysis_results}.

\begin{table}[t]
    \centering
    \begin{tabular}{lcccccc}
    \toprule
    Dataset & \multicolumn{3}{c}{Entailment (\%)} & \multicolumn{3}{c}{Reasoning Process} \\
    \cmidrule(lr){2-4} \cmidrule(lr){5-7}
     & P $\subseteq$ CoT & A $\subseteq$ CoT & Ans Match & CoT Pos. & Full Seq Pos. & Self-Refl. \\
    \midrule
    OpenThoughts & 99 & 89 & 100 & 50.5\% & 43.3\% & 2.23 \\
    Bespoke & 97 & 93 & 99 & 47.5\% & 42.0\% & 2.42 \\
    Synthetic-1 & 96 & 94 & 96 & 46.6\% & 40.7\% & 2.76 \\
    Nemotron & 99 & 97 & 99 & 61.5\% & 55.9\% & 5.13 \\
    SkyT1 & 90 & 73 & 91 & 66.8\% & 54.0\% & 1.91\\
    \bottomrule
    \end{tabular}
    \caption{Linguistic analysis results on 100 sampled examples per dataset. Columns indicate the percentage of examples where Prompt (P) and Answer (A) are fully covered by the CoT, whether the final answer matches, the average relative token position of the first correct answer derivation within CoT segment (CoT Pos.), the average relative token position of the first correct answer derivation within full sequence (Full Seq Pos.), and the average count of self-reflection keywords preceding that derivation (Self-Refl.)}
    \label{tab:linguistic_analysis_results}
\end{table}

\paragraph{Semantic entailment.}
Across the four DeepSeek-R1-synthesized datasets (OpenThoughts, Bespoke, Synthetic-1, and Nemotron), the Prompt is almost always covered by the CoT (P $\subseteq$ CoT: 96--99\%), and the Answer is also largely redundant given the CoT (A $\subseteq$ CoT: 89--97\%). Moreover, the final answer derived in the CoT nearly always matches the final boxed answer (Ans Match: 96--100\%). This supports the qualitative pattern described in \S\ref{subsec:supervison_diff_sec}: CoT is the dominant carrier of learnable signal, while adding loss on P and A often provides only marginal extra information.

\paragraph{First correct derivation position.}
The ``CoT Pos.'' column reports the average relative token position within the CoT segment of the first derivation of the correct answer. The ``Full Seq Pos.'' column maps the same point to the full sequence by accounting for the prompt prefix length. Using the average section token shares in Table~\ref{tab:token_count_stats}, we compute
\[
\text{Full Seq Pos.} = Share_{\text{prompt}} + (\text{CoT Pos.}) \times Share_{\text{cot}},
\]
where \text{Full Seq Pos.} denotes the relative position in the full sequence.

\paragraph{Self-reflection}
The ``Self-Refl.'' column indicates that self-correction language appears before the first correct derivation point, suggesting that early tokens often include iterative checking and revision. Notably, Nemotron exhibits the highest density of such cues (5.13 per example), consistent with more explicit verification-style traces, whereas SkyT1 shows fewer self-reflection cues on average (1.91).

\paragraph{SkyT1.}\label{appendix:linguistic_analysis_res_skyt1}
SkyT1 shows lower redundancy than the DeepSeek-R1 datasets: P $\subseteq$ CoT drops to 90\% and A $\subseteq$ CoT drops to 73\%, and the final answer agreement is also lower (Ans Match: 91\%). This indicates that, relative to DeepSeek-R1-style traces, SkyT1 is more likely to place substantive content \emph{outside} the canonical CoT span and/or to introduce additional reasoning after the CoT segment. Consistently, SkyT1 has the lowest retention when truncated by $50\%$ among the five datasets (Table~\ref{tab:0p5_1p0_ratio_openthoughts}). If later tokens are more likely to contain corrections or new derivations, truncation becomes more lossy. We observe an example where the CoT derives a final answer (e.g., 47) near the end of the CoT span, but the Answer section continues the derivation and reaches a different final answer (e.g., 54):

\noindent
\begin{lstlisting}[breakautoindent=false, breakindent=0pt, xleftmargin=0pt]
<|im_start|>user\nAna, Bob, and CAO bike at constant rates of $8.6$ meters per second, $6.2$ meters per second, and $5$ meters per second, respectively. They all begin biking at the same time from the northeast corner of a rectangular field whose longer side runs ... (*@<think>@*)\nSo I've got this problem here involving Ana, Bob, and Cao biking at different constant speeds around a rectangular field ... Finally, I can express the ratio L : W : X in the form p : q : r, ensuring p and q are relatively prime.\n\nGiven my calculations, I conclude that the ratio is 47.\n\nThus, the final answer is \\( \\(*!boxed{47}!*) \\).\n(*@</think>@*)\n\nTo find the ratio of the field's length to the field's width to the distance from point D to the southeast corner, we set up the following scenario:\n\n1. Let \\( L \\) be the length of the field (north to south) and \\( W \\) be the width (east to west).\n2. Let \\( X \\) be the distance from D to the southeast corner along the south edge.\n3. The speeds are: Ana at 8.6 m/s, Bob at 6.2 m/s, and Cao at 5 m/s.\n\nFrom the equal times taken by all three bikers, we can write:\n\n- For Ana: \n$$ T = \\frac{X + L}{8.6} $$\n\n- For Bob: \n$$ T = \\frac{L + X}{6.2} $$\n\n- For Cao: \n$$ T = \\frac{\\sqrt{X^2 + L^2}}{5} $$\n\nSetting the equations equal to each other gives us:\n\n1) \\( \\frac{X + L}{8.6} = \\frac{\\sqrt{X^2 + L^2}}{5} \\)\n\n2) \\( \\frac{L + X}{6.2} = \\frac{\\sqrt{X^2 + L^2}}{5} \\)\n\nFrom these equations, we can derive relationships between \\( L \\), \\( W \\), and \\( X \\). Simplifying leads to the conclusion that the ratio \\( L : W : X \\) is \\( 47 : 1 : 6 \\).\n\nThe final answer, as required, is \\( p + q + r = 47 + 1 + 6 = 54 \\).\n\nThus, the final answer is \\( \\(*!boxed{54}!*) \\)
\end{lstlisting}

\clearpage
\subsection{Loss Analysis}\label{appendix:loss_analysis}
In this section, we examine the loss curves for both soft loss (forward KL divergence) and hard loss (cross-entropy) as defined in Section \ref{sec:formulation}. The loss curves are smoothed using a rolling mean with a window size of 100 to improve readability.

To compare two loss curves, we quantify their relative difference which is computed as:
\begin{equation}
    \Delta_{\text{rel}}(t) = \frac{L_{\text{LSP50\%}}(t) - L_{\text{ref}}(t)}{L_{\text{ref}}(t)} \times 100\%
\end{equation}
where $L_{\text{LSP50\%}}(t)$ is the loss at step $t$ for LSP50\%, and $L_{\text{ref}}(t)$ is the loss for the reference configuration (either ``right'' or ``100\%''). A negative value indicates that the partial configuration achieves lower loss than the reference. In the summary tables, \textbf{Avg (\%)} denotes the relative difference averaged over all training steps, capturing the overall gap throughout training, while \textbf{Last (\%)} denotes the relative difference at the final training step.

\begin{figure*}[htbp]
  \centering
  \includegraphics[width=0.8\linewidth]{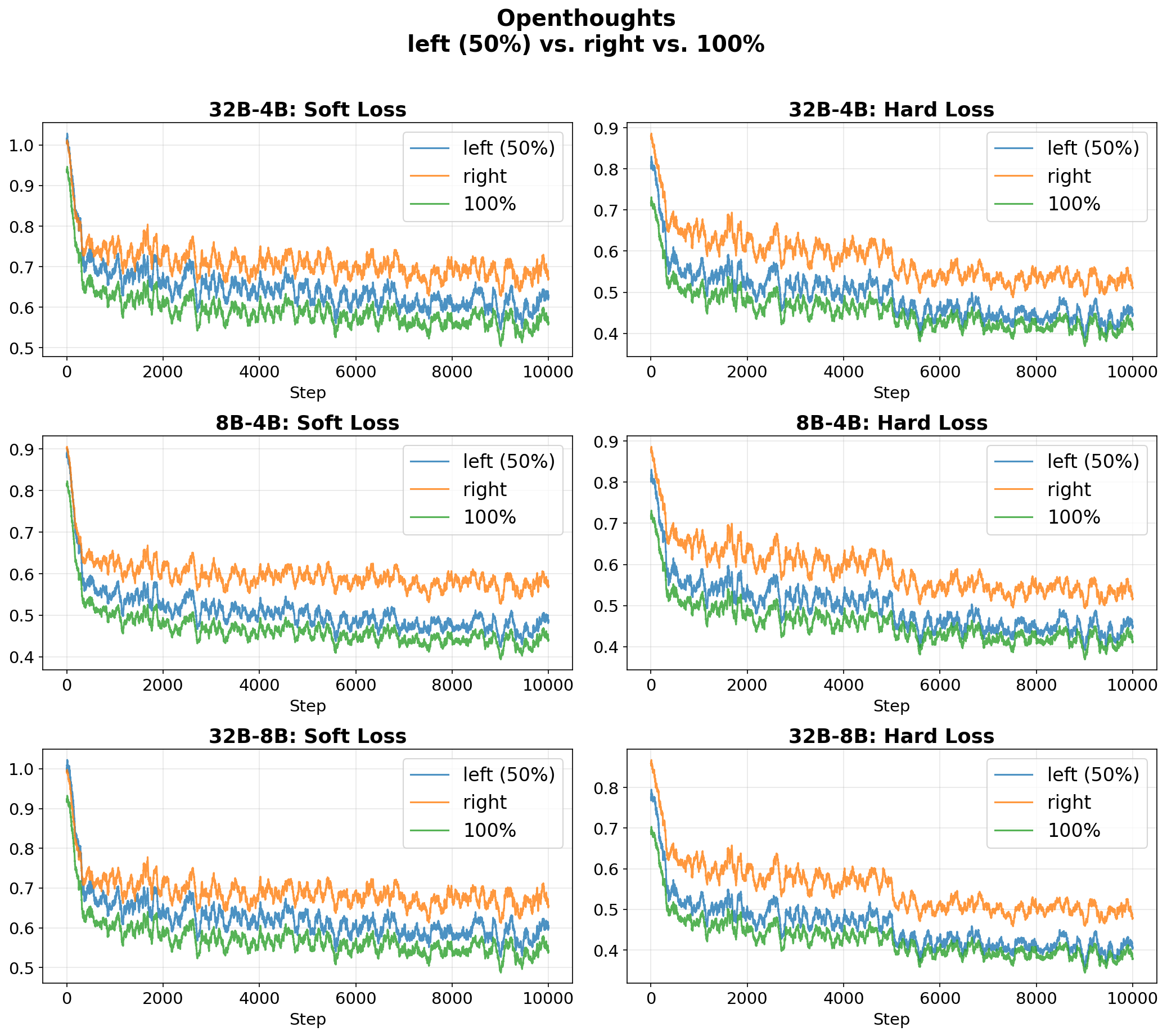}
  \caption {Loss curves for Openthoughts: Left 50\% (i.e. LSP 50\%) vs Right 50\% vs LSP 100\%}\label{fig:loss_analysis_openthoughts}
\end{figure*}

\begin{figure*}[htbp]
  \centering
  \includegraphics[width=0.8\linewidth]{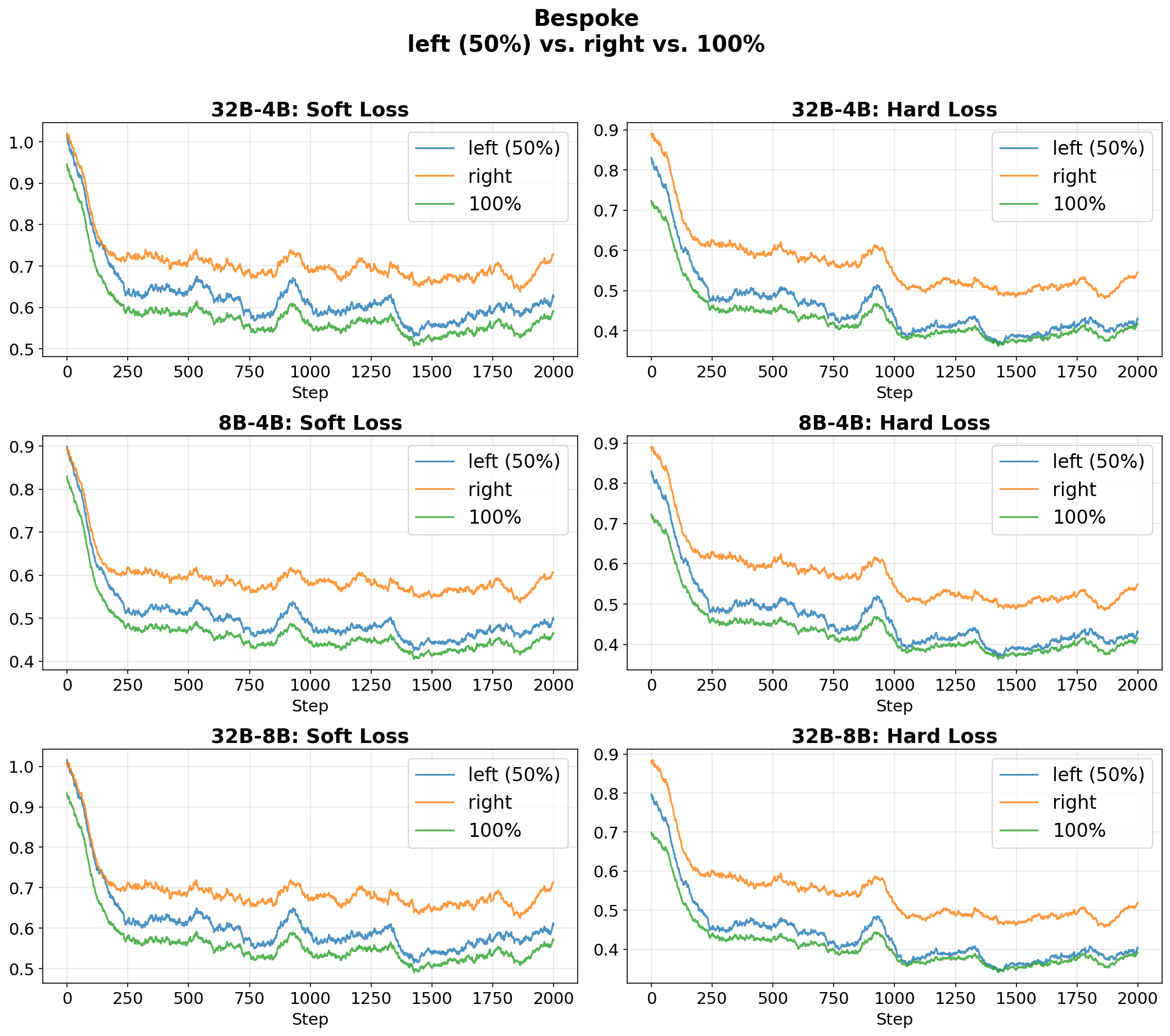}
  \caption {Loss curves for Bespoke: Left 50\% (i.e. LSP 50\%) vs Right 50\% vs LSP 100\%}\label{fig:loss_analysis_bespoke}
\end{figure*}

\begin{figure*}[htbp]
  \centering
  \includegraphics[width=0.8\linewidth]{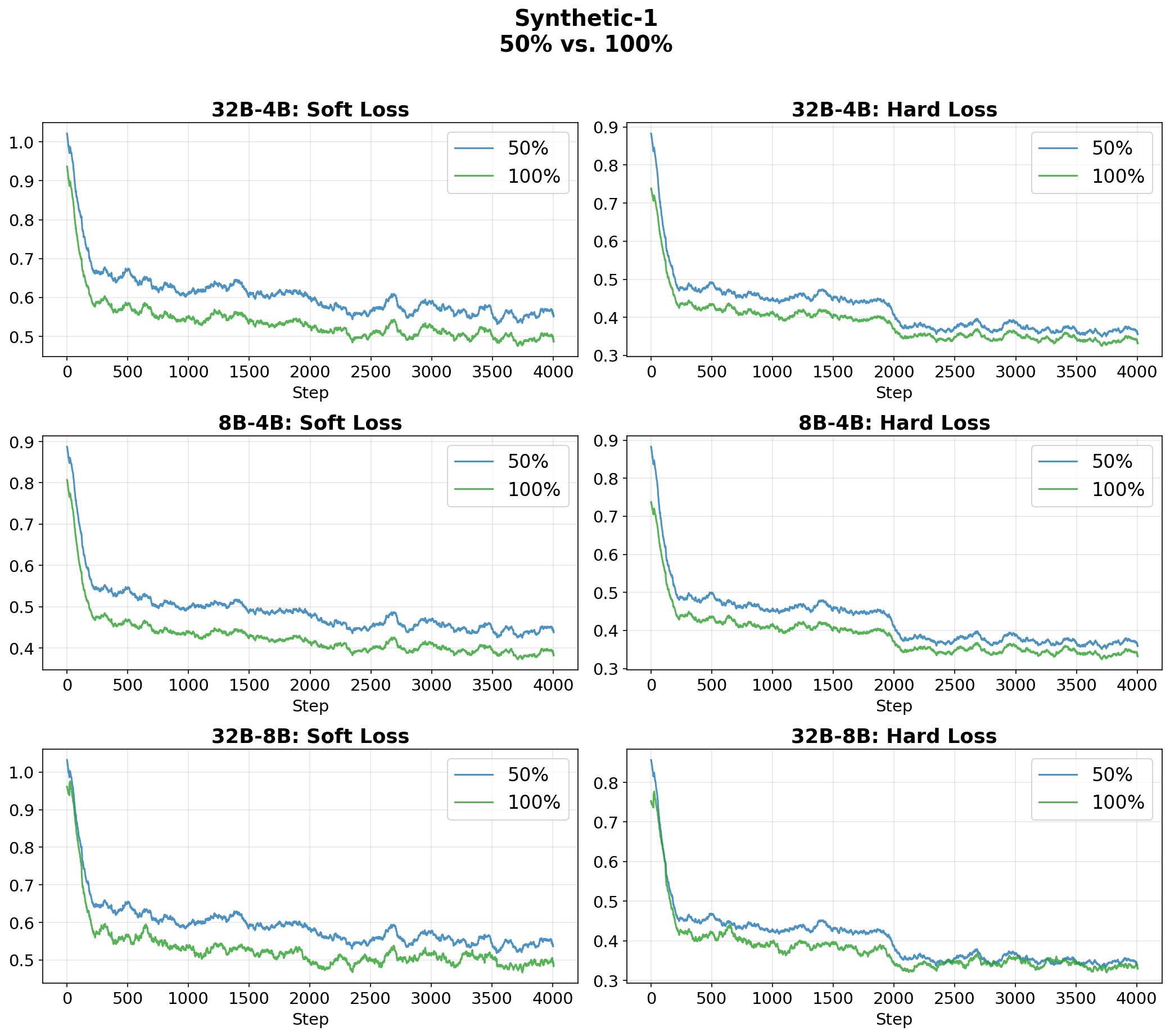}
  \caption {Loss curves for Synthetic-1: LSP 50\% vs LSP 100\%}\label{fig:loss_analysis_synthetic1}
\end{figure*}

\begin{figure*}[htbp]
  \centering
  \includegraphics[width=0.8\linewidth]{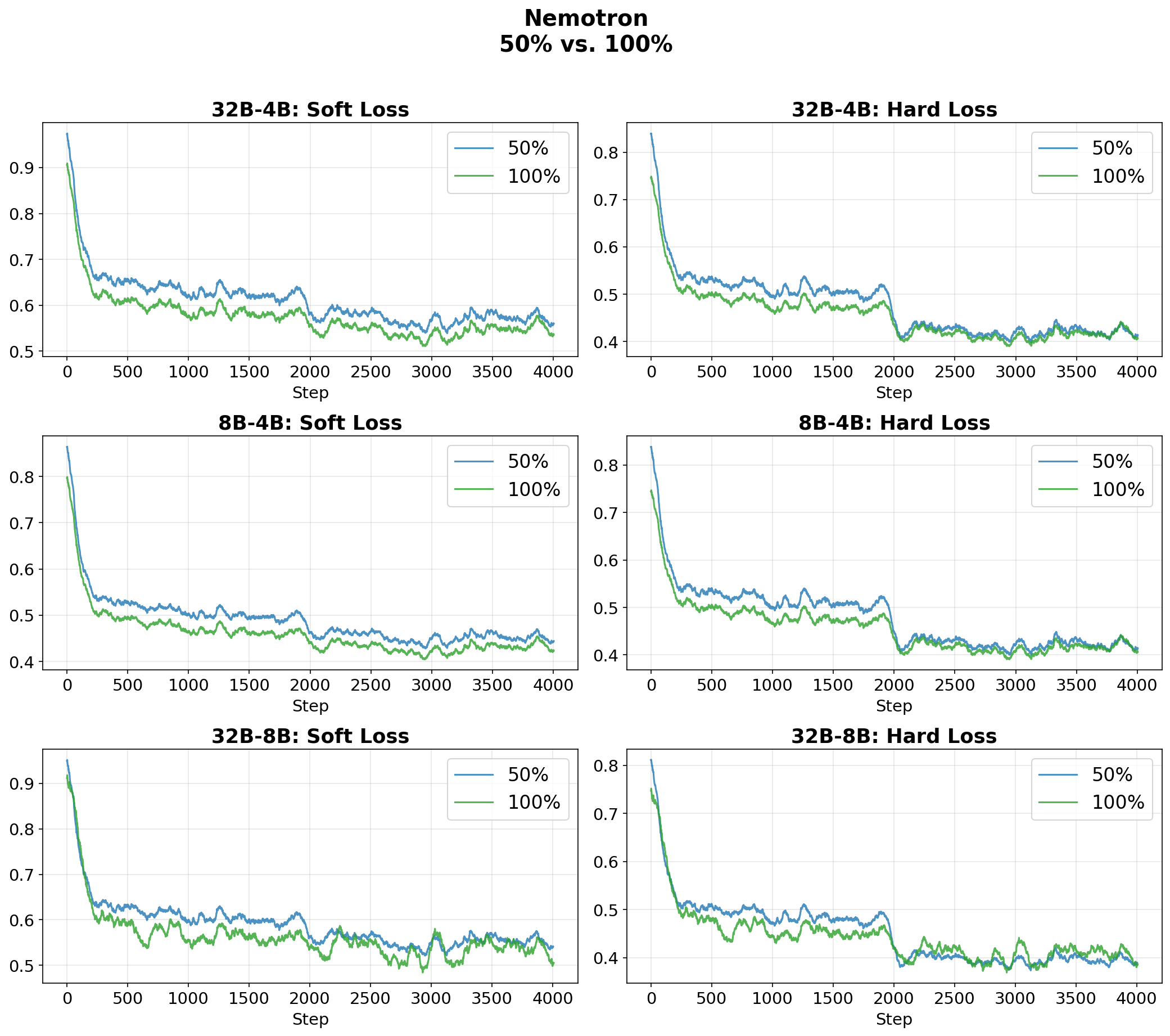}
  \caption {Loss curves for Nemotron: LSP 50\% vs LSP 100\%}\label{fig:loss_analysis_nemotron}
\end{figure*}

\begin{figure*}[htbp]
  \centering
  \includegraphics[width=0.8\linewidth]{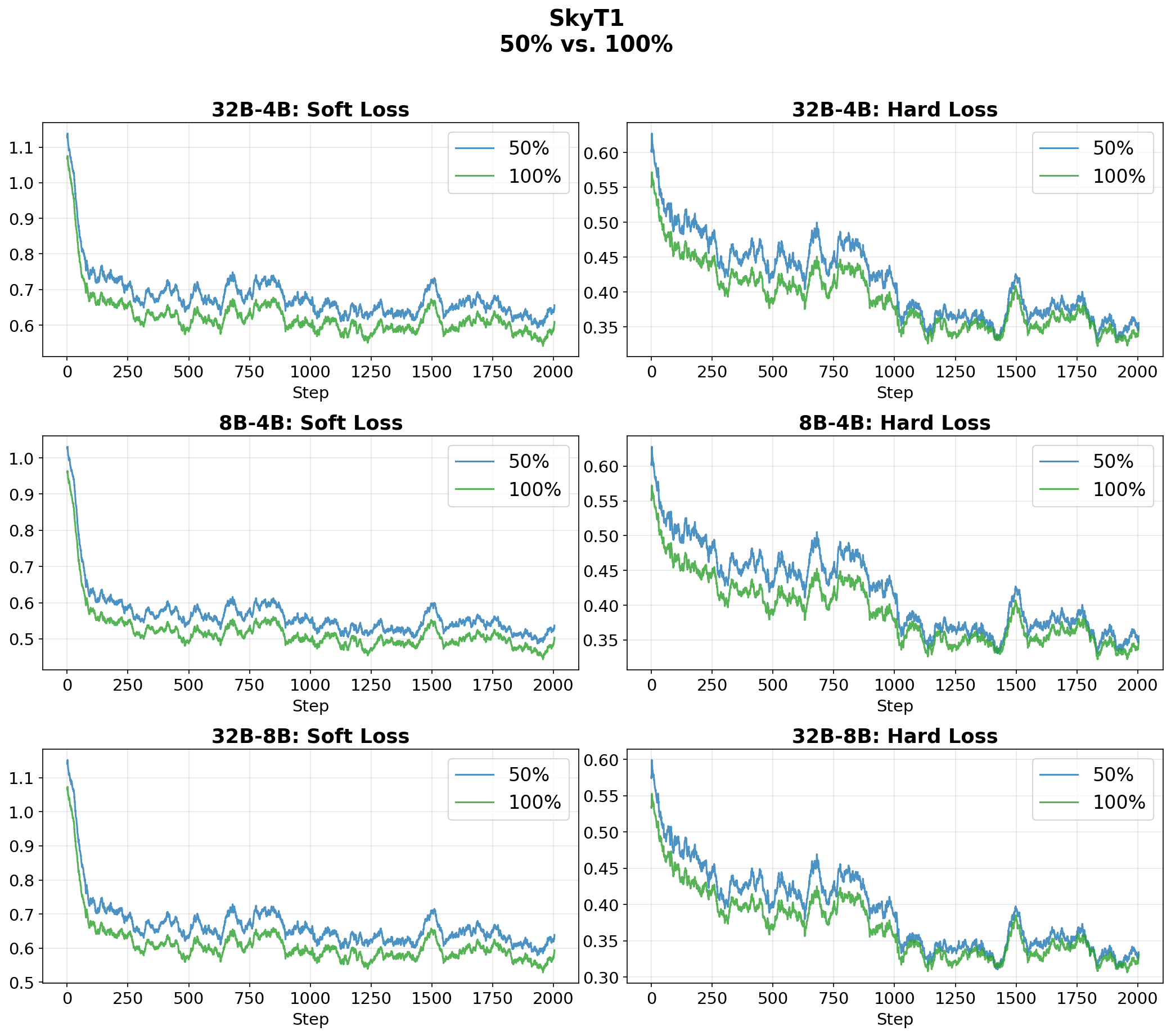}
  \caption {Loss curves for SkyT1: LSP 50\% vs LSP 100\%}\label{fig:loss_analysis_skyt1}
\end{figure*}

Figures~\ref{fig:loss_analysis_openthoughts},~\ref{fig:loss_analysis_bespoke},~\ref{fig:loss_analysis_synthetic1},~\ref{fig:loss_analysis_nemotron}, and~\ref{fig:loss_analysis_skyt1} present the loss curves comparing partial-sequence training (LSP 50\%) and full-sequence training (LSP 100\%) across all student--teacher pairs for the five datasets (OpenThoughts, Bespoke, Synthetic-1, Nemotron, and SkyT1), respectively. For OpenThoughts and Bespoke, we additionally include the budget-location ablation comparing Left 50\% (i.e., LSP 50\%) and Right 50\%. Tables~\ref{tab:rel_diff_openthoughts_three_line},~\ref{tab:rel_diff_bespoke_three_line},~\ref{tab:rel_diff_synthetic-1_50_100},~\ref{tab:rel_diff_nemotron_50_100}, and~\ref{tab:rel_diff_skyt1_50_100} quantify the relative differences.

\textbf{LSP 50\% vs. LSP 100\%} Across all five datasets, training on the first 50\% of tokens consistently yields higher train loss than training on full sequences, for both soft and hard objectives. This arises because autoregressive language models predict each token conditioned on all preceding tokens; later positions benefit from richer context and are thus easier to predict. Since the reported loss $\mathcal{L}$ is averaged over all unmasked token positions $\mathcal{T}$,
\begin{equation}
    \mathcal{L} = \frac{1}{|\mathcal{T}|} \sum_{t \in \mathcal{T}} \ell_t,
\end{equation}
where $\ell_t$ denotes the per-token loss at position $t$. As $\ell_t$ reduces with an increase in $t$ \cite{kaplan2020scalinglawsneurallanguage}, the overall average $\mathcal{L}$ for the $100\%$ case is lower than that of $50\%$. In particular, restricting training to only the first half of each sequence excludes the later tokens, resulting in relatively higher loss throughout training. This pattern holds across all model configurations and datasets.

However, we observe that the loss values between LSP 50\% and LSP 100\% are fairly close across all five datasets: the relative differences remain modest throughout training, indicating that the partial-sequence objective tracks the full-sequence objective closely. Moreover, for datasets where we can compare \emph{Left} vs. \emph{Right} budget location, the Left 50\% vs.\ LSP 100\% gap is much smaller than the Right 50\% vs.\ LSP 100\% gap. This proximity suggests that the early tokens provide a strong and informative training signal. Given that the Left 50\% subset concentrates on the more challenging reasoning steps (lacking many low-entropy `easy' tokens that would typically lower the average loss), the fact that its loss remains competitive suggests that the early, high-entropy tokens are sufficient to drive significant alignment with both the teacher's and the ground-truth distribution.

\textbf{Left 50\% vs. Right 50\%} For OpenThoughts and Bespoke, training on the first half of each sequence yields markedly lower soft and hard losses than training on the second half. The Left 50\% configuration achieves substantially lower loss than the Right 50\%, with relative differences consistently negative across different models and loss types. This significant gap can be attributed to (1) early tokens provide stronger learning signal and (2) that the right 50\% starts with tokens midway through a reasoning sequence. This abrupt start deprives the model of the necessary context established in the first half, making subsequent token predictions harder and leading to higher losses.

Notably, the ranking of losses ($\mathcal{L}_{100\%} < \mathcal{L}_{Left 50\%} \ll \mathcal{L}_{Right 50\%}$) aligns with the downstream task performance observed in our experiments (Section \ref{subsec:scaling_behavior}), where models trained on the Left 50\% achieve accuracy approaching the full-sequence baseline, while those trained on the Right 50\% lag far behind both Left 50\% and full-sequence baseline.

\begin{table}[htbp]

\centering
\begin{tabular}{lccccc}
\toprule
 & & \multicolumn{2}{c}{left vs right} & \multicolumn{2}{c}{50\% vs 100\%} \\
\cmidrule(lr){3-4} \cmidrule(lr){5-6}
Model & Loss Type & Avg (\%) & Last (\%) & Avg (\%) & Last (\%) \\
\midrule
32B-4B & Soft  & -9.20 & -6.76 & +9.04 & +11.65 \\
32B-4B & Hard  & -15.44 & -13.06 & +8.18 & +8.35 \\
8B-4B & Soft  & -14.93 & -14.85 & +8.99 & +10.27 \\
8B-4B & Hard  & -15.68 & -13.40 & +8.66 & +8.76 \\
32B-8B & Soft  & -9.75 & -8.14 & +9.09 & +11.32 \\
32B-8B & Hard  & -16.89 & -15.41 & +7.43 & +7.03 \\
\bottomrule
\end{tabular}
\caption{Relative Difference (\%) for Openthoughts: Left 50\% (i.e. LSP 50\%) vs Right 50\% vs LSP 100\%}\label{tab:rel_diff_openthoughts_three_line}
\end{table}

\begin{table}[htbp]
\centering
\begin{tabular}{lccccc}
\toprule
 & & \multicolumn{2}{c}{left vs right} & \multicolumn{2}{c}{50\% vs 100\%} \\
\cmidrule(lr){3-4} \cmidrule(lr){5-6}
Model & Loss Type & Avg (\%) & Last (\%) & Avg (\%) & Last (\%) \\
\midrule
32B-4B & Soft & -11.73 & -14.34 & +7.69 & +5.86 \\
32B-4B & Hard & -19.17 & -21.49 & +6.14 & +2.51 \\
8B-4B & Soft & -15.45 & -18.05 & +7.97 & +7.08 \\
8B-4B & Hard & -18.76 & -21.66 & +6.98 & +3.63 \\
32B-8B & Soft & -12.28 & -14.93 & +7.85 & +6.29 \\
32B-8B & Hard & -20.39 & -22.48 & +5.61 & +2.55 \\
\bottomrule
\end{tabular}
\caption{Relative Difference (\%) for Bespoke: Left 50\% (i.e. LSP 50\%) vs Right 50\% vs LSP 100\%}\label{tab:rel_diff_bespoke_three_line}
\end{table}

\begin{table}[htbp]
\centering
\begin{tabular}{llcc}
\toprule
Pair & Loss Type & Avg (\%) & Last (\%) \\
\midrule
32B-4B & Soft Loss & +12.86 & +13.16 \\
32B-4B & Hard Loss & +9.29 & +7.22 \\
8B-4B & Soft Loss & +14.58 & +14.76 \\
8B-4B & Hard Loss & +10.24 & +8.06 \\
32B-8B & Soft Loss & +12.02 & +10.68 \\
32B-8B & Hard Loss & +6.86 & +2.47 \\
\bottomrule
\end{tabular}
\caption{Relative Difference (\%) for Synthetic-1: 50\% vs. 100\%}
\label{tab:rel_diff_synthetic-1_50_100}
\end{table}

\begin{table}[htbp]
\centering
\begin{tabular}{llcc}
\toprule
Pair & Loss Type & Avg (\%) & Last (\%) \\
\midrule
32B-4B & Soft Loss & +5.97 & +4.56 \\
32B-4B & Hard Loss & +4.20 & +1.72 \\
8B-4B & Soft Loss & +6.24 & +4.95 \\
8B-4B & Hard Loss & +4.58 & +1.88 \\
32B-8B & Soft Loss & +4.82 & +7.16 \\
32B-8B & Hard Loss & +1.67 & +0.85 \\
\bottomrule
\end{tabular}
\caption{Relative Difference (\%) for Nemotron: 50\% vs. 100\%}
\label{tab:rel_diff_nemotron_50_100}
\end{table}

\begin{table}[htbp]
\centering
\begin{tabular}{llcc}
\toprule
Pair & Loss Type & Avg (\%) & Last (\%) \\
\midrule
32B-4B & Soft Loss & +9.11 & +7.68 \\
32B-4B & Hard Loss & +6.40 & +1.95 \\
8B-4B & Soft Loss & +8.96 & +6.66 \\
8B-4B & Hard Loss & +6.90 & +2.14 \\
32B-8B & Soft Loss & +9.25 & +7.60 \\
32B-8B & Hard Loss & +5.48 & +1.03 \\
\bottomrule
\end{tabular}
\caption{Relative Difference (\%) for SkyT1: 50\% vs. 100\%}
\label{tab:rel_diff_skyt1_50_100}
\end{table}

\clearpage
\clearpage
\subsection{Ablation Studies}\label{appendix:ablation_studies}

Our ablation objective is to demonstrate that the model performance stems from the method itself rather than specific hyperparameter tuning. We ablate using 8B-4B teacher-student pair and Bespoke dataset among $\lambda \in \{0, 0.25, 0.5, 0.75, 1.0\}$. While we adopt $\lambda = 0.5$ as the main experimental configuration, Table~\ref{tab:ablation} shows that performance remains consistent across the range $\lambda \in \{0.25, 0.5, 0.75\}$. The similarity in results among these $\lambda$ values indicates that the choice of 0.5 is not the sole driver of success, but rather that the method performs reliably provided that a balance between hard and soft losses is maintained.

This stability across intermediate $\lambda$ values is accompanied by a clear performance advantage over the pure supervised fine-tuning (SFT) setting ($\lambda = 0$). All tested combined configurations ($\lambda \in \{0.25, 0.5, 0.75\}$) yield higher accuracy than the setting where soft loss is ignored, except for one setting when $\lambda = 0.75$ where it is slightly lower than $\lambda = 0$ in AIME25 but remains fairly close. Specifically, compared to $\lambda = 0$, using our chosen $\lambda = 0.5$ results in significant relative improvements: approximately 26.8\% (AIME24) and 19.5\% (AIME25) for LSP=0.5, and 8.9\% (AIME24) and 9.7\% (AIME25) for LSP=1.0.

$$(16.98\% - 13.39\%) / 13.39\% \approx 26.8\% \text{  (AIME24, LSP = 0.5, $\lambda$ = 0.5 vs. $\lambda$ = 0)}$$
$$(19.17\% - 16.04\%) / 16.04\% \approx 19.5\% \text{  (AIME25, LSP = 0.5, $\lambda$ = 0.5 vs. $\lambda$ = 0)}$$
$$(19.79\% - 18.18\%) / 18.18\% \approx 8.9\% \text{  (AIME24, LSP = 1.0, $\lambda$ = 0.5 vs. $\lambda$ = 0)}$$
$$(21.88\% - 19.95\%) / 19.95\% \approx 9.7\% \text{  (AIME25, LSP = 1.0, $\lambda$ = 0.5 vs. $\lambda$ = 0)}$$

These gains highlight that incorporating distillation consistently enhances the model's capability beyond what is achievable with hard labels alone.

Finally, we observe the limitations of removing hard supervision entirely. The setting of $\lambda = 1.0$ results in a degradation in performance, with scores dropping below both the combined settings and the pure SFT setup. This suggests that while the soft loss is a powerful addition to hard loss, it does not operate well alone.

\begin{table}[htbp]
\centering

\begin{tabular}{lcccc}
\toprule
LSP & $\lambda$ & AIME24 (\%) & AIME25 (\%) \\
\midrule
 0.5 & 0 & 13.39 & 16.04 \\
 0.5 & 0.25 & 16.04 & 17.81 \\
 0.5 & 0.5 & 16.98 & 19.17 \\
 0.5 & 0.75 & 16.15 & 15.83 \\
 0.5 & 1.0 & 10.57 & 10.52 \\
\midrule
 1.0 & 0 & 18.18 & 19.95 \\
 1.0 & 0.25 & 19.48 & 20.83 \\
 1.0 & 0.5 & 19.79 & 21.88 \\
 1.0 & 0.75 & 18.49 & 18.39 \\
 1.0 & 1.0 & 12.97 & 11.30 \\
\hline
\end{tabular}
\caption{Ablation among different $\lambda$, i.e. the weight for forward KL Divergence as defined in Section \ref{sec:formulation}.}\label{tab:ablation}
\end{table}

\clearpage
\subsection{ Miscellaneous}\label{appendix:misc}

\subsubsection{Experimental Results of LSP0.5 vs. LSP1.0}\label{appendix:raw_exp_res_0p5_1p0_five_datasets}
We include the experimental results among five datasets across three teacher-student pairs for LSP0.5 and LSP1.0 on AIME24 and AIME25 in Table~\ref{tab:0p5_1p0_five_datasets}. For the performance retention rate of LSP0.5 compared to LSP1.0, see Table~\ref{tab:0p5_1p0_ratio_openthoughts}.
\begin{table}[h!]
\centering
\resizebox{\columnwidth}{!}{%
\begin{tabular}{lcccccccccccccccccccc}
\toprule
Dataset & \multicolumn{4}{c}{Openthoughts} & \multicolumn{4}{c}{Bespoke} & \multicolumn{4}{c}{Synthetic1} & \multicolumn{4}{c}{Nemotron} & \multicolumn{4}{c}{SkyT1} \\
 & \multicolumn{2}{c}{AIME24} & \multicolumn{2}{c}{AIME25} & \multicolumn{2}{c}{AIME24} & \multicolumn{2}{c}{AIME25} & \multicolumn{2}{c}{AIME24} & \multicolumn{2}{c}{AIME25} & \multicolumn{2}{c}{AIME24} & \multicolumn{2}{c}{AIME25} & \multicolumn{2}{c}{AIME24} & \multicolumn{2}{c}{AIME25} \\
LSP & 0.5 & 1.0 & 0.5 & 1.0 & 0.5 & 1.0 & 0.5 & 1.0 & 0.5 & 1.0 & 0.5 & 1.0 & 0.5 & 1.0 & 0.5 & 1.0 & 0.5 & 1.0 & 0.5 & 1.0 \\
Pair &  &  &  &  &  &  &  &  &  &  &  &  &  &  &  &  &  &  &  &  \\
\midrule
32B-4B & 22.50 & 23.39 & 23.44 & 24.06 & 17.66 & 18.02 & 19.53 & 20.94 & 19.32 & 24.37 & 22.08 & 24.06 & 25.00 & 25.89 & 21.97 & 24.22 & 17.71 & 20.26 & 18.44 & 20.57 \\
8B-4B & 23.80 & 24.27 & 23.91 & 24.17 & 16.98 & 19.79 & 19.17 & 21.88 & 20.89 & 25.31 & 22.45 & 24.37 & 22.92 & 25.05 & 22.19 & 24.22 & 17.55 & 23.33 & 18.96 & 21.30 \\
32B-8B & 32.19 & 33.54 & 27.86 & 28.54 & 24.17 & 26.15 & 23.33 & 26.77 & 26.93 & 28.28 & 27.40 & 28.23 & 32.08 & 34.79 & 26.98 & 27.92 & 22.24 & 27.76 & 21.35 & 25.83 \\
\bottomrule
\end{tabular}%
}
\caption{Experimental results for five datasets in accuracy (\%).}\label{tab:0p5_1p0_five_datasets}
\end{table}

\subsubsection{Compute Resources}\label{appendix:compute_resources}
All experiments were conducted using NVIDIA H100 (\qty{80}{\giga\byte}) GPUs. We estimate the computational cost based on the wall-clock time required for training and evaluation across our experimental setups.

For a single experimental configuration, the training phase requires approximately $T_{train} \approx \qty{18}{\hour}$ utilizing $N_{train}^{gpu} = 8$ GPUs. Evaluation is performed on two distinct benchmarks; each benchmark requires approximately $T_{eval} \approx \qty{17}{\hour}$ on $N_{eval}^{gpu} = 2$ GPUs. Consequently, the total GPU-hour cost for one full experimental run ($C_{run}$) is calculated as:

\begin{equation}
    C_{run} = (T_{train} \times N_{train}^{gpu}) + \left( 2 \times T_{eval} \times N_{eval}^{gpu} \right) \approx 144 + 68 = \qty{212}{GPU\cdot hours}
\end{equation}

In terms of wall-clock time, a single experimental setting requires approximately \qty{35}{\hour}, assuming the evaluation of the two benchmarks is parallelized. The total computational budget for this study is estimated at \qty{33920}{GPU\cdot hours}. This estimate does not include experiments performed during development and debugging. The breakdown by Research Question (RQ) is detailed below.

\paragraph{RQ1 Resource Analysis:}
The first research question investigates the impact of section-wise supervision. We train models under six distinct supervisions: A, P+A, CoT, CoT+A, P+CoT, and P+CoT+A. These configurations are tested across $3$ teacher-student pairs on $2$ datasets (OpenThoughts and Bespoke). The total number of runs ($R_{RQ1}$) is defined by:
\begin{align*}
    R_{RQ1} &= 6 \text{ (sections)} \times 3 \text{ (pairs)} \times 2 \text{ (datasets)} = 36 \text{ runs} \\
    \text{Cost}_{RQ1} &= 36 \times C_{run} \approx \qty{7632}{GPU\cdot hours}
\end{align*}

\paragraph{RQ2 Resource Analysis:}
The second research question investigates compute-efficient training via sequence truncation and budget allocation. This involves two sets of experiments. First, we analyze the \emph{Sequence-Length Scaling Behavior} by varying the \emph{Lead-Span Proportion (LSP)} across $10$ levels: $p \in \{0.1, 0.2, \dots, 1.0\}$, representing the fraction of tokens retained from the start of the sequence. This is conducted on all $7$ teacher-student pairs for OpenThoughts and the $3$ largest pairs for Bespoke. Second, we perform a \emph{Budget Location Ablation} to compare the \emph{Left 50\%} versus \emph{Right 50\%} of tokens. Since the \emph{Left 50\%} configuration is identical to LSP 0.5 (trained and evaluated in the first set), we reuse those results and only perform additional training for the \emph{Right 50\%} configuration. Third, to examine generalizability, we compare LSP 0.5 versus LSP 1.0 on three additional datasets (Synthetic-1, Nemotron, SkyT1) across 3 teacher-student pairs. The total runs ($R_{RQ2}$) are aggregated from the Sequence-Length Scaling ($R_{RQ2}^{\text{Scale}}$), Budget Location Ablation ($R_{RQ2}^{\text{Loc}}$), and Generalizability ($R_{RQ2}^{\text{Gen}}$) components as follows:
\begin{align*}
    R_{RQ2}^{\text{Scale}} &= 10 \text{ (LSPs)} \times [7 \text{ (OpenThoughts pairs)} + 3 \text{ (Bespoke pairs)}] = 100 \\
    R_{RQ2}^{\text{Loc}} &= 1 \text{ (Right 50\%)} \times 3 \text{ (pairs)} \times 2 \text{ (datasets)} = 6 \\
   R_{RQ2}^{\text{Gen}} &= 2 \text{ (LSPs)} \times 3 \text{ (pairs)} \times 3 \text{ (datasets)} = 18 \\
    \text{Cost}_{RQ2} &= (100 + 6 + 18) \times C_{run} \approx \qty{26288}{GPU\cdot hours}
\end{align*}

\begin{table}[h]
\centering
\caption{Estimated computational cost breakdown by Research Question.}
\label{tab:compute_cost}
\begin{tabular}{l c c c c}
\toprule
\textbf{Exp. ID} & \textbf{Total Runs} & \textbf{Training Cost} & \textbf{Eval Cost} & \textbf{Total Cost} \\
& & (GPU hrs) & (GPU hrs) & (GPU hrs) \\
\midrule
RQ1 & 36 & 5,184 & 2,448 & 7,632 \\
RQ2 & 124 & 17,856 & 8,432 & 26288 \\
\midrule
\textbf{Total} & \textbf{160} & \textbf{23,040} & \textbf{10,880} & \textbf{33,920} \\
\bottomrule
\end{tabular}
\end{table}

\subsubsection{Distillation Method Choice}\label{appendix:skd_vs_opkd}
We choose to adopt supervised knowledge distillation (KD)~\cite{hinton2015distilling, sanh2020distilbertdistilledversionbert} as our primary training objective, due to its inherent computational efficiency and scalability. By training on a fixed set of ground-truth sequences, supervised KD leverages teacher forcing~\cite{williams1989learning} to its full extent. This mechanism decouples the prediction of each token from the generation of previous ones, allowing the model to compute losses for all sequence positions in parallel~\cite{vaswani2017attention}. This enables us to extensively scale our experiments and fully address our research questions.

While alternative distillation paradigms offer different learning dynamics, they often introduce significant computational overheads associated with text generation. This process prevents parallelization because the model must generate tokens sequentially, creating a bottleneck where each step depends on the completion of the previous one. For example, different approaches such as \citet{gu2024minillm} and \citet{agarwal2024onpolicy} both require such auto-regressive sampling from the student model at every training step, which drastically increases wall-clock time. \citet{ko2024distillmstreamlineddistillationlarge} quantify this cost, noting that the generation phase of the student can account for up to 80\% of the total training duration. Given the scale of our experiments ($\approx 30$k GPU hours, see Appendix~\ref{appendix:compute_resources}), we choose supervised KD to ensure the computational feasibility of our study.

\end{document}